%% file: main.tex
\documentclass[final,3p,times]{elsarticle}

\usepackage{amssymb}

\usepackage{algorithmic}
\usepackage[ruled,vlined]{algorithm2e}

\usepackage{multirow} 
\usepackage{booktabs}
\usepackage{adjustbox}

\usepackage{xcolor}
\usepackage{color}

\usepackage{amsmath}

\begin{document}

\begin{frontmatter}



\title{Word-Sequence Entropy: Towards Uncertainty Estimation in Free-Form Medical Question Answering Applications and Beyond}


\author[inst1]{Zhiyuan Wang}
\author[inst2]{Jinhao Duan}
\author[inst3]{Chenxi Yuan}
\author[inst4]{Qingyu Chen}
\author[inst5]{Tianlong Chen}
\author[inst2]{Yue Zhang}
\author[inst6]{Ren Wang}
\author[inst1]{\\Xiaoshuang Shi\corref{corr1}}
\ead{xsshi2013@gmail.com}
\cortext[corr1]{Corresponding author}
\author[inst2]{Kaidi Xu}

\affiliation[inst1]{organization={Center for Future Media, School of Computer Science and Engineering, University of Electronic Science and Technology of China},
            city={Chengdu}, 
            state={Sichuan},
            postcode={611731},
            country={China}}

\affiliation[inst2]{organization={Department of Computer Science, Drexel University},
            city={Philadelphia},
            state={PA},
            postcode={19104},
            country={USA}}

\affiliation[inst3]{organization={Department of Biostatistics, Epidemiology, and Informatics (DBEI), Perelman School of Medicine, University of Pennsylvania},
            city={Philadelphia}, 
            state={PA},
            postcode={19104},
            country={USA}}

\affiliation[inst4]{organization={Section of Biomedical Informatics $\&$ Data Science, Yale School of Medicine, Yale University},
            city={New Haven},
            state={CT},
            postcode={06510}, 
            country={USA}}

\affiliation[inst5]{organization={Computer Science $\&$ Artificial Intelligence Laboratory, Massachusetts Institute of Technology},
            city={Cambridge}, 
            state={MA}, 
            postcode={02139}, 
            country={USA}}

\affiliation[inst6]{organization={Department of Electrical and Computer Engineering, Illinois Institute of Technology},
            city={Chicago}, 
            state={IL},
            postcode={60616},
            country={USA}}

\begin{abstract}
Uncertainty estimation is crucial for the reliability of safety-critical human and artificial intelligence (AI) interaction systems, particularly in the domain of healthcare engineering. 
However, a robust and general uncertainty measure for free-form answers has not been well-established in open-ended medical question-answering (QA) tasks, where generative inequality introduces a large number of irrelevant words and sequences within the generated set for uncertainty quantification (UQ), which can lead to biases. 
This paper introduces Word-Sequence Entropy (\textit{WSE}), a method that calibrates uncertainty at both the word and sequence levels, considering semantic relevance. \textit{WSE} quantifies uncertainty in a way that is more closely aligned with the reliability of LLMs during uncertainty quantification (UQ). 
We compare \textit{WSE} with six baseline methods on five free-form medical QA datasets, utilizing seven popular large language models (LLMs). Experimental results demonstrate that \textit{WSE} exhibits superior performance in UQ under two standard criteria for correctness evaluation.
Additionally, in terms of real-world medical QA applications, the performance of LLMs is significantly enhanced (e.g., a 6.36\% improvement in model accuracy on the COVID-QA dataset) by employing responses with lower uncertainty that are identified by \textit{WSE} as final answers, without any additional task-specific fine-tuning or architectural modifications. 
\end{abstract}


\end{frontmatter}


\input{section/introduction}
\input{section/related_work}
\input{section/proposed_method}
\input{section/experiment}
\input{section/conclusion}

\section*{Acknowledgement}
Zhiyuan Wang and Xiaoshuang Shi were supported by the National Key Research $\&$ Development Program of China under Grant (No. 2022YFA1004100). 



\bibliographystyle{elsarticle-num} 
\bibliography{cas-refs}





\end{document}

%% file: section/introduction.tex
\section{Introduction}
\label{sec: intro}
Healthcare professionals and patients increasingly employ online search engines to query information and symptoms when confronted with medical conditions. 
A U.S. health survey~\cite{abacha2015means} found that $18\%$ of individuals who self-diagnosed online received conflicting advice or outright refusals from medical experts. 
Despite this, about $77\%$ of adults still prefer online searches over in-person consultations, posing significant health risks. 
In this context, there is a pressing demand for reliable question-answering (QA) applications in healthcare, to provide accurate and trustworthy responses to user queries. 

Recent advancements in natural language generation (NLG), particularly in question-answering (QA)~\cite{brown2020language,chowdhery2022palm,chen2023chatcot,ouyang2022training}, have been driven by large language models (LLMs)~\cite{waisberg2023gpt,zhang2022opt,touvron2023llama,he2023can}. 
Enabled by in-context learning\footnote{In-context learning is to design task-specific instruction prompts, and then leverage a few annotated samples as the prompts to guide LLMs to tackle new test data.} (ICL)~\cite{min2021noisy}, LLMs exhibit outstanding task-agnostic and few-shot performance~\cite{brown2020language,chowdhery2022palm,duan2024gtbench}. 
Given a few-shot prompt with multiple query-response pairs, LLMs efficiently handle new QA tasks~\cite{brown2020language,ouyang2022training}, showing great potential for real-world medical QA applications. 
However, LLMs are proven to \textit{hallucinate}\footnote{``Hallucinate'' is defined as LLMs generating content that is nonsensical or unfaithful to the provided source content. 
In this case, users cannot trust that any output is correct.} and provide unfactual answers that seem plausible but deviate from user instructions~\cite{manakul2023selfcheckgpt,yao2024survey,sun2024trustllm,hongdecoding}, compromising the reliability of their deployment in healthcare applications. 
Uncertainty quantification (UQ) is an effective approach to address these issues~\cite{kadavath2022language, chen2023quantifying}. 
By estimating the uncertainty of statements, practical QA applications can inform users about the trustworthiness of the query-answering process, thereby mitigating the risk of unforeseen health incidents.

Nevertheless, UQ in free-form QA tasks, particularly in the medical domain, poses significant challenges. 
Unlike prediction tasks with specific output forms and labels~\cite{yuan2023remind}, LLMs-based QA generate semantically equivalent responses but syntactically or lexically distinct, resulting in an unbounded output space. 
Additionally, LLMs face multiple sources of uncertainty, primarily aleatoric uncertainty from data distribution and epistemic uncertainty from insufficient information~\cite{kendall2017uncertainties}. 
To address these issues, existing methods either empower LLMs to self-evaluate the uncertainty of their answers through fine-tuning~\cite{lin2022teaching,kadavath2022language} or devise entropy-based measures~\cite{malinin2020uncertainty,kuhn2023semantic,duan2023shifting}. 
Recent work, Shift Attention to Relevance ($\textit{SAR}$)~\cite{duan2023shifting}, reallocates the weights of uncertainty induced by each token and sentence based on their relevance, achieving state-of-the-art performance in multiple general-purpose QA tasks. 

\begin{figure*}[!t]
\centerline{\includegraphics[width=1.0\columnwidth]{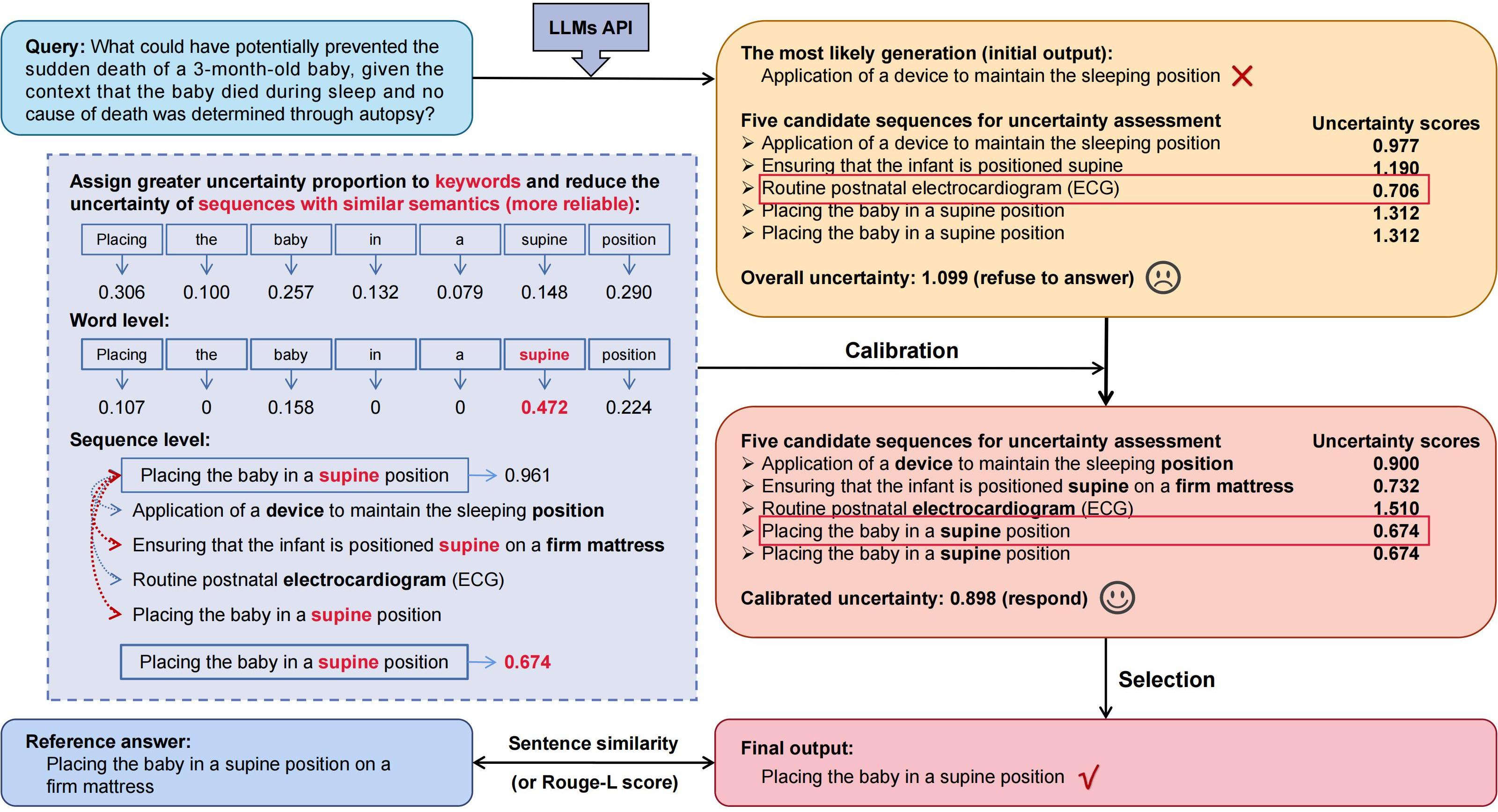}}
\caption{The overview of \textit{WSE} and its potential for improving model accuracy. 
Given a medical query, the language model generates the most likely generation as the output, which might be incorrect. 
Following prior work, we additionally generate multiple (e.g., five) candidate responses to evaluate the trustworthiness of this output. 
Existing entropy-based measures identify high overall uncertainty in the candidate set, causing the API to refuse to answer the most likely generation. 
By assessing semantic relevance at both the word and sequence levels, \textit{WSE} highlights keywords and reliable sequences, resulting in calibrated uncertainty that meets the response criterion. 
Finally, we employ the response with the lowest uncertainty as the final output, which coincides with the reference answer.}
\label{fig: pipeline}
\end{figure*}

In open-ended medical QA tasks, a general framework for quantifying the uncertainty of free-form responses has yet to be established. 
An overview of our method is illustrated in Fig.~\ref{fig: pipeline}. 
Generative inequality introduces many irrelevant words and sequences within the candidate responses for UQ, leading to biased uncertainty measurements when existing entropy-based methods treat all words and sequences equally. 
To address this issue, we propose $\textbf{W}$ord-$\textbf{S}$equence $\textbf{E}$ntropy ($\textit{WSE}$), which allocates greater uncertainty proportion to relevant components, e.g., tokens and sentences, making the estimated uncertainty more well-aligned to the semantics of generations.
Additionally, we leverage the concept of bi-directional entailment \cite{kuhn2023semantic}—if two textual sequences logically imply each other, they are semantically similar—to develop a new method for measuring the semantic textual similarity between two sequences, which correlates with semantic relevance. 
Moreover, we investigate improving model accuracy by resampling based on the uncertainty measure, aiming to mitigate the limitations of LLMs in the medical domain.

We evaluate $\textit{WSE}$ utilizing multiple open-source pre-trained (e.g., LLaMA-7B~\cite{touvron2023llama}) and instruction-tuned (e.g., LLaMA-2-7B-Chat~\cite{touvron2023llama2}, StableBeluga-7B~\cite{touvron2023llama2,mukherjee2023orca} and Zephyr-7B-Alpha~\cite{tunstall2023zephyr}) LLMs with the model size of 7B on five open-ended medical QA datasets (i.g., COVID-QA~\cite{moller2020covid}, Medical Meadow MedQA~\cite{jin2020disease}, PubMedQA~\cite{jin2019pubmedqa}, MedMCQA~\cite{pal2022medmcqa} and MedQuAD~\cite{ben2019question}). 
Experimental results show that \textit{WSE} outperforms six baseline methods (e.g., \textit{WSE} surpasses \textit{SAR} by 4.99\% AUROC on the PubMedQA dataset). 
Furthermore, after filtering sequences with high uncertainty identified by \textit{WSE}, we obtain a substantial improvement in model accuracy (e.g., +6.36\% accuracy on the COVID-QA dataset, utilizing the Zephyr-7B-Alpha model), demonstrating the remarkable potential in real-world medical QA applications. 

Our major contributions are summarized as follows:

\begin{itemize}
    \item We investigate the phenomenon of generative inequality within the responses generated by LLMs in open-ended medical QA tasks and analyze its implications for uncertainty measurement.
    \item We propose Word-Sequence Entropy ($\textit{WSE}$) to quantify the uncertainty of free-form answers in open-ended medical QA tasks for the first time. 
    \item We conduct extensive experiments on five free-form medical QA datasets utilizing seven LLMs under two standard criteria for correctness evaluation, demonstrating that \textit{WSE} surpasses six comparable baselines. 
    \item Without requiring additional task-specific fine-tuning or architectural modifications, we improve the performance of LLMs, by resampling and applying responses with lower uncertainty, measured by \textit{WSE}, as final answers, and obtain remarkable enhancement of model accuracy.
\end{itemize}

%% file: section/related_work.tex
\section{Related Work}
\label{sec: related works}
\subsection{UQ in Conventional NLP Tasks}
The concepts and approaches of UQ have been extensively explored and analyzed across various tasks~\cite{hullermeier2021aleatoric}, including machine translation (MT). 
To address data uncertainty from semantically equivalent translations, under-specification, and lower-quality training data in MT, Ott et al.~\cite{ott2018analyzing} assess whether references match the top model prediction or if most generated sequences align well with human translations. 
Considering the relationship between model probabilities and human judgments, Fomicheva et al.~\cite{fomicheva2020unsupervised} establish a strong correlation with human quality judgments through UQ techniques. 
Glushkova et al.\cite{glushkova2021uncertainty} address accumulated uncertainty from noisy scores, insufficient references, and out-of-domain text by incorporating Monte Carlo (MC) dropout~\cite{gal2016dropout} and model ensembling~\cite{lakshminarayanan2017simple}, characterizing uncertainty through confidence intervals. 

Due to limited work on calibration in a regression setting, Wang et al.~\cite{wang2022uncertainty} augment training data in low-resource scenarios and select instances based on UQ, addressing both the data and model predictive uncertainty. 
Malinin et al.~\cite{malinin2020regression} also apply prior networks for interpretable UQ.

To enhance the reliability of decision-making in text classification tasks, Miok et al.~\cite{miok2019prediction} quantify the predictive uncertainty utilizing MC dropout regularization~\cite{gal2016dropout} and detect hate speech efficiently and reliably. 
Given the fundamental notion of epistemic uncertainty (EU)~\cite{kendall2017uncertainties} as a lack of knowledge, Lahlou et al.~\cite{lahlou2021deup} introduce the approach of direct epistemic uncertainty prediction (DEUP) and assess the excess risk as a measure of EU. 

\subsection{UQ in Free-form NLG Tasks}
Distinguishing tasks with specific labels, such as misclassification detection~\cite{vazhentsev2022uncertainty} and text classification~\cite{hu2021uncertainty}, it is challenging to implement uncertainty estimation in open-ended NLG tasks, where any output from LLMs sharing equivalent semantics with the standard answer can be considered correct. 

The issue of truthfulness motivates uncertainty calibration for LLMs. 
Lin et al.~\cite{lin2022teaching} empower LLMs to self-evaluate the uncertainty of their answers in words via supervised fine-tuning. 
Meanwhile, Kadavath et al.~\cite{kadavath2022language} adopt answer options from existing multiple-choice tasks and ask LLMs to determine if each answer is true or false. Both approaches prompt the language model itself to measure uncertainty with additional task-specific training. 
In a zero-resource setting, Manakul et al.~\cite{manakul2023selfcheckgpt} attribute poor performance to variations in generating patterns. If the consistency score of multiple generations is low, it indicates high uncertainty. 
Motivated by the limited work on general uncertainty estimation for structured prediction, Malinin et al.~\cite{malinin2020uncertainty} devise a novel measure of knowledge uncertainty by summing the predictive entropy over multiple outputs. 
Recently, to tackle the issue of semantic equivalence, Kuhn et al.~\cite{kuhn2023semantic} propose to cluster semantically similar sequences and calculate the semantic entropy. 
The approach most connected with ours is \textit{SAR}~\cite{duan2023shifting}, which reassigns the weight of uncertainty associated with each token and sentence based on their respective relevance.

Compared to general-purpose QA tasks, medical QA with free-form responses is more domain-specific and often involves rare and compound technical terms. 
In such cases, LLMs adopt character-based tokenization, breaking a single word into multiple sub-tokens for processing.  
Analyzing each token independently, as done in \textit{SAR}, can lead to inconsistency of semantic relevance within the same word, resulting in biased and unstable uncertainty measurements. 
Additionally, \textit{SAR} relies on an external language model to measure semantic similarity, which lacks explainability and reliability due to the semantic complexity of medical QA. 
Given the absence of a robust and general approach to estimating uncertainty in open-ended medical QA tasks, we aim to address this gap by developing an estimator to inform users about the trustworthiness of output statements from LLMs.

%% file: section/proposed_method.tex
\section{Methodology}
\label{sec: proposed method}

\subsection{Preliminaries}
\label{sec: preliminaries}
Conditioned on a medical query $x$, LLMs progressively predict the probability distribution of the next token based on previous tokens and generate free-form textual sequences in an auto-regressive fashion. 
Following prior work~\cite{kuhn2023semantic, duan2023shifting}, we generate $K$ responses to the same query and estimate the predictive uncertainty of the current QA process within the generated set $\mathbb{S}=\{\textbf{s}_1, \textbf{s}_2, \cdots, \textbf{s}_K\}$, where $\textbf{s}_i$ refers to the $i$-th response. 
We denote the $j$-th word within the textual sequence ${\textbf{s}}_{i}$ as ${w}_{ij}$, and the $k$-th token in ${w}_{ij}$ as ${z}_{ijk}$. 
Additionally, we denote the number of words within ${\textbf{s}}_{i}$ by ${N}_{i}$, the number of tokens in ${w}_{ij}$ by ${M}_{j}$ and the total number of tokens within ${\textbf{s}}_{i}$ by ${T}_{i}$ (i.e., ${T}_{i} = \textstyle\sum_{j}^{{N}_{i}} {M}_{j}$). 
Prompted by $x$, we define the probability of generating ${z}_{t}$ as $p\left ({z}_{t} \mid {\textbf{z}}_{<t}, x \right)$, where ${\textbf{z}}_{<t}\left (t\in {T}_{i} \right )$ refers to previously generated tokens within the $i$-th textual sequence. 
In subsequent research, we simplify $p\left ({z}_{t} \mid {\textbf{z}}_{<t}, x \right)$ to $p\left ({z}_{t} \right)$ to represent the generative probability of the $t$-th token. 

\subsection{Generative Inequality in Free-form Medical Query Responses}
\label{sec: generative inequality}
To investigate the issue of generative inequality in open-ended medical QA tasks, we leverage the popular Predictive Entropy ($\textit{PE}$)~\cite{kadavath2022language} as the fundamental method for UQ. 
Given ${\textbf{s}}_{i}$, we first calculate the token-wise entropy of ${z}_{t}$ based on its generative probability:
\begin{equation}\label{eq: token-wise entropy}
{\textit{E}}_{T}\left ( {z}_{t}\right )=-\log p\left ({z}_{t} \right).
\end{equation}
Then, \textit{PE} calculates the sequence-wise entropy of $\textbf{s}_i$ by summing the per-token entropy:
\begin{equation}\label{eq: sequence-wise entropy 1}
{\textit{E}}_{S}\left ({\textbf{s}}_{i} \right )=\displaystyle\sum_{t}^{{T}_{i}}{\textit{E}}_{T}\left ( {z}_{t}\right ) .
\end{equation}
The predictive uncertainty or entropy of the current QA process is obtained by averaging the sequence-wise entropy of these $K$ candidate responses: 
\begin{equation}\label{eq: predictive entropy}
\textit{E}\left (\mathbb{S} \right )=\frac{1}{K}\displaystyle\sum_{i}^{K}{\textit{E}}_{S}\left ({\textbf{s}}_{i} \right ) .
\end{equation}

In this context, it is apparent that the token-wise entropy represents the uncertainty committed by individual tokens, the sequence-wise entropy captures the predictive uncertainty of each textual sequence (i.e., response), and \textit{PE} quantifies the complexity encompassing the generated set (i.e., an approximation of the model's output space), which characterizes the overall uncertainty of the current decision-making process for medical queries. 

Analogous to the formulation of token-wise entropy in Eq.~\eqref{eq: token-wise entropy}, the sequence-wise entropy of $\textbf{s}_i$ can be expressed as its log-probability:
\begin{equation} \label{eq: sequence-wise entropy 2}
{\textit{E}}_{S}\left ({\textbf{s}}_{i} \right )=-\log p\left ( {\textbf{s}}_{i} \right ),
\end{equation}
where $p\left ( {\textbf{s}}_{i} \right )$ reflects the probability of the $i$-th sequence and is obtained by multiplying the probabilities of all tokens within ${\textbf{s}}_{i}$ (i.e., $\textstyle\prod_{t}^{{T}_{i}}p\left ({z}_{t} \right )$).

\subsubsection{Relevance}
\label{sec: relevance}
To analyze generative inequality at the word level, where keywords (e.g., ``Mother-to-child transmission'' in the sentence ``Mother-to-child transmission is the primary cause of HIV-1 infection in children worldwide.'') may account for a limited proportion of the overall uncertainty within the current response, we first assess the semantic relevance of each word by measuring the textual similarity between the query-answer pairs before and after removing the evaluated word. 
A lower similarity score signifies a significant semantic variation, indicating that the word carries more semantic information within the current textual sequence (i.e., a keyword). 

Following \textit{SAR}~\cite{duan2023shifting}, we evaluate textual similarity utilizing a cross-encoder model provided by the SentenceTransformers library~\cite{reimers2019sentence}, with RoBERTa-large~\cite{liu2019roberta} as the backbone. 
The model processes sentence pairs and generates similarity scores.  
However, relying solely on an external language model for textual similarity evaluation is unreliable and lacks explainability, because embeddings of sentences encoded by the model, in which all semantic information is mixed in fixed-length vectors, are limited in the semantic representation~\cite{wang2023going}. 
Inspired by bi-directional entailment~\cite{kuhn2023semantic}, we leverage a Natural Language Inference (NLI) classifier, DeBERTa-large-mnli~\cite{he2020deberta}, for this task. 
The model takes sequence pairs as the input and predicts scores (logits) for three classes of semantic relationship: entailment, neutral, and contradiction. 
We employ the probability of entailment as the similarity measure.

For simplicity, we define ${\textbf{s}}_{i} \setminus {w}_{ij}$ as the representation for removing the $j$-th word from the $i$-th response and $\cup$ as the concatenation of the prompt and answer. 
The measurement of textual similarity is formulated as: 
\begin{equation}\label{eq: semantic similarity score}
\left\{
\begin{array}{l}
     {S}_{C}={f}_{\textit{ce}}\left (x \cup {\textbf{s}}_{i} , x \cup {\textbf{s}}_{i} \setminus {w}_{ij} \right) \\
     {S}_{L}= {f}_{\textit{ent}}\left (x \cup {\textbf{s}}_{i} , x \cup {\textbf{s}}_{i} \setminus {w}_{ij} , c \right),
\end{array}
\right.
\end{equation}
where ${f}_{\textit{ce}}\left (\cdot  \right )$ represents the utilization of the \textit{cross-encoder} model to compute the textual similarity score between two sequences directly, ${f}_{\textit{ent}}\left (\cdot  \right )$ refers to obtaining the probability of \textit{entailment} extracted from the logit vector, which falls within the range of 0 to 1 after being scaled by the $softmax$ function, and $c$ is leveraged to control the smoothness of the logit vector.

Given that the employed language models~\cite{liu2019roberta, he2020deberta} are not specifically pre-trained for the medical domain, consistently high similarity can lead to low semantic relevance for all words within the current textual sequence, thereby failing to capture keywords. 
We adopt a conservative strategy by selecting the smaller value from the two measures in Eq.~\eqref{eq: semantic similarity score}, which mitigates potential instability arising from extreme similarity quantification and task-specific limitations. 
Then, the word-level semantic relevance score of the $j$-th word within the $i$-th response can be formulated as:
\begin{equation}\label{eq: word-level semantic relevance score}
{\textit{R}}_{W}\left ( {w}_{ij} \right ) = 1 - \min \left ( {S}_{C}, {S}_{L}\right ) .
\end{equation}
In the end, we assign the same relevance score to all tokens in ${w}_{ij}$ as the word itself (i.e., the token-level semantic relevance score), to maintain the consistency of semantic relevance within a single word: 
\begin{equation}\label{eq: token-level semantic relevance score}
{\textit{R}}_{T}\left ( {z}_{ijk} \right ) = {\textit{R}}_{W}\left ( {w}_{ij} \right ) \left ( k\in M_j\right ).
\end{equation}
Formally, it can be observed that if the $i$-th textual sequence exhibits significant semantic variation before and after removing the $j$-th word, then the semantic relevance score of all tokens in ${w}_{ij}$ are deemed to be high.

In open-ended medical QA tasks, we generate multiple (i.e., $K$) responses to the same query to estimate the uncertainty of the current QA process, and there can be many irrelevant responses with limited semantic information. 
However, \textit{PE}, as described in Eq.~\eqref{eq: predictive entropy}, calculates the average of the sequence-wise entropy of all responses within the generated set. 
To investigate this issue, we define the semantic relevance at the sequence level. 

Building on the self-consistency hypothesis\footnote{Self-consistency hypothesis states that a repetitively sampled response is viewed as a form of consistency linked to higher confidence in the response.}~\cite{wang2022self}, we suggest that responses, which maintain strong semantic consistency with others among the set of $K$ candidate responses, are more trustworthy. 
We employ the identical approaches described in Eq.~\eqref{eq: semantic similarity score} to measure the textual similarity between any two textual sequences. 
Then, the sequence-level relevance score of ${\textbf{s}}_{i}$ is formulated as the accumulation of the textual similarity scores, re-weighted by the generative probability of the compared responses:
\begin{equation}\label{eq: sequence-level semantic relevance score}
{\textit{R}}_{S}\left ( {\textbf{s}}_{i} \right )=\displaystyle\sum_{l\neq i}^{K} S\left ({\textbf{s}}_{l},{\textbf{s}}_{i} \right )p\left ( {\textbf{s}}_{l} \right  ) ,
\end{equation}
where $S \left (\cdot,\cdot  \right )$ represents the smaller similarity score obtained from the two measurements in Eq.~\eqref{eq: semantic similarity score}, and $\textbf{s}_l$ denotes the $l$-th textual sequence that differs from $\textbf{s}_i$ in the $K$ generated responses. 
A higher probability of $\textbf{s}_l$ (i.e., $p\left ( {\textbf{s}}_{l} \right  )$) augments the persuasiveness of textual similarity between ${\textbf{s}}_{i}$ and ${\textbf{s}}_{l}$.

\subsubsection{Uncertainty}
As mentioned previously, the token-wise entropy reflects the uncertainty committed by each token (i.e., ${\textit{E}}_{T}\left ( {z}_{t}\right )$ in Eq.~\eqref{eq: token-wise entropy}), and the overall uncertainty of the $i$-th response can be calculated by aggregating the token-wise entropy of all words within the entire textual sequence (i.e., ${\textit{E}}_{S}\left ({\textbf{s}}_{i} \right )$ in Eq.~\eqref{eq: sequence-wise entropy 1}). 
To ascertain how much uncertainty is induced by individual words, we compute the word-wise entropy of ${w}_{ij}$ based on Eq.~\eqref{eq: token-wise entropy}:
\begin{equation}\label{eq: word-wise entropy}
{\textit{E}}_{W}\left ({w}_{ij} \right )=\displaystyle\sum_{k}^{{M}_{j}}-\log p\left ({z}_{ijk} \right),
\end{equation} 
where $p\left ({z}_{ijk} \right)$ refers to the probability of generating ${z}_{ijk}$ as the $k$-th token in the $j$-th word within the $i$-th response. 
Then we calculate the ratio of the word-wise entropy and the sequence-wise entropy to determine the proportion of uncertainty stemming from the $j$-th word within the $i$-th response.:
\begin{equation}\label{eq: word-wise uncertainty proportion}
{\textit{P}}_{W}\left ({w}_{ij},{\textbf{s}}_{i} \right )=\frac{{\textit{E}}_{W}\left ( {w}_{ij}\right )}{{\textit{E}}_{S}\left ( {\textbf{s}}_{i}\right )}.
\end{equation}

Similar to the word-wise situation, we formulate the uncertainty proportion of the $i$-th response in the set of $K$ generated responses (i.e., $\mathbb{S}$) as:
\begin{equation}\label{eq: sequence-wise uncertainty proportion}
{\textit{P}}_{S}\left ({\textbf{s}}_{i},\mathbb{S} \right )=\frac{{\textit{E}}_{S}\left ( {\textbf{s}}_{i}\right )}{\textstyle\sum_{l}^{K}{\textit{E}}_{S}\left ( {\textbf{s}}_{l}\right )}.
\end{equation}

\begin{figure}[!t]
\centerline{\includegraphics[width=0.55\columnwidth]{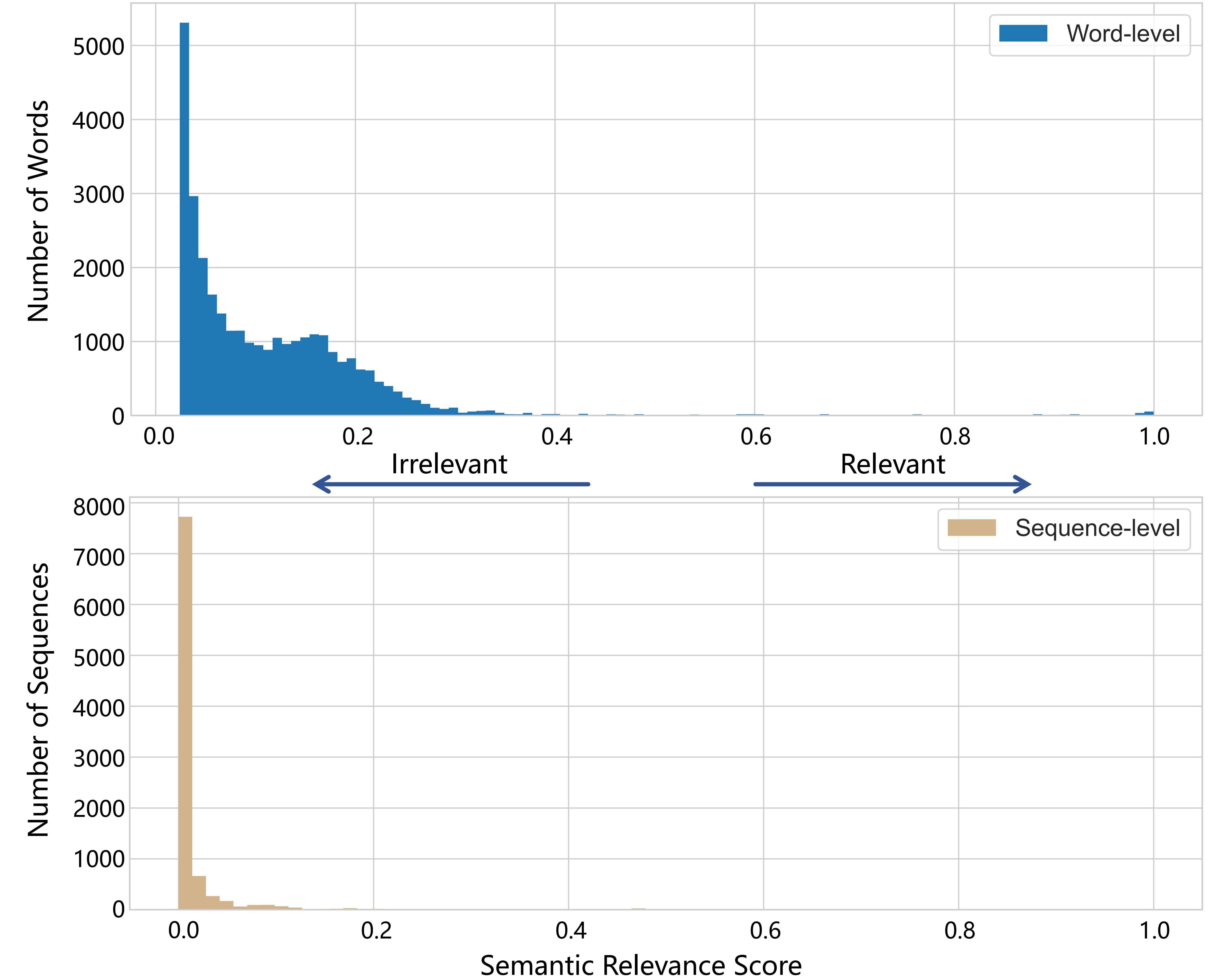}}
\caption{Distribution of semantic relevance scores at both the word and sequence levels. The entire generated set contains a considerable proportion of irrelevant words and sequences (i.e., generative inequality).}
\label{fig: relevance distribution}
\end{figure}

\begin{figure}[!t]
\centerline{\includegraphics[width=0.55\columnwidth]{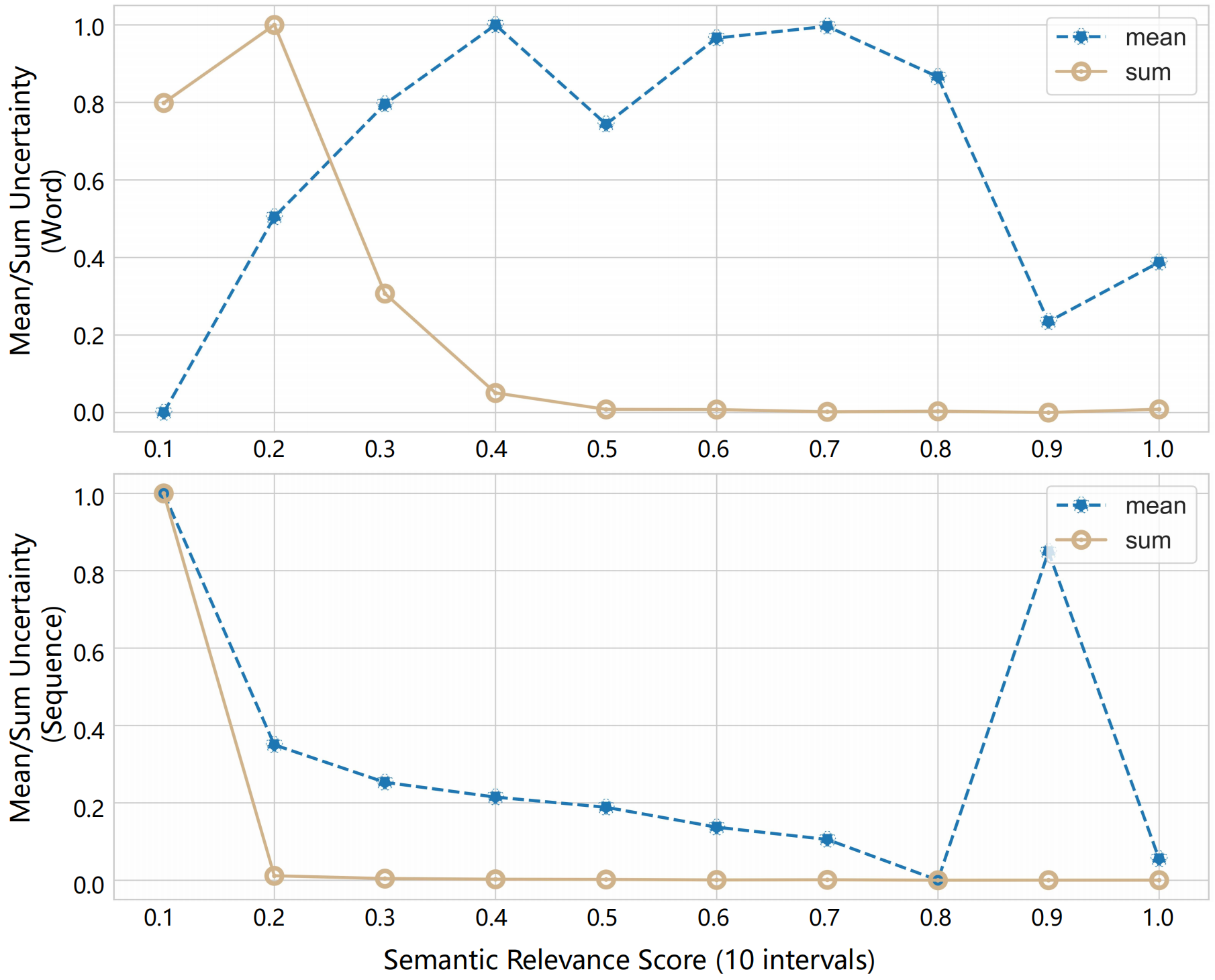}}
\caption{Correlation between the semantic relevance and uncertainty proportion at both the word and sequence levels. 
Irrelevant words and sequences account for the primary source of uncertainty within the generated set (responses) in general.}
\label{fig: correlation}
\end{figure}

\subsubsection{Correlation Analysis}
\label{sec: verification}
To characterize generative inequality in open-ended medical QA tasks, we employ the MedMCQA dataset, with LLaMA-2-7B-Chat-HF serving as the generator. 
Given each medical query, we generate five responses (i.e., $K=5$), and the max length of each sequence is set to 128 (i.e., ${T}_{i}\leqslant 128$). 
We first leverage Eq.~\eqref{eq: word-level semantic relevance score} and Eq.~\eqref{eq: token-level semantic relevance score} to outline the distributions of word-level and sequence-level semantic relevance scores. 
Results are depicted in Fig.~\ref{fig: relevance distribution}. 
Within the generated set, a considerable proportion of words exhibit low semantic relevance (i.e., irrelevant), and only a limited subset of words conveys the primary semantic information. 
At the sequence level, the prevalence of irrelevant responses significantly outweighs those with meaningful content. 

When conducting UQ, we should prioritize keywords and reliable textual sequences. 
To explore the issue of generative inequality, we analyze the correlation between semantic relevance and uncertainty proportion. 
We divide relevance scores into ten equal intervals.  
Within each interval, we calculate the sum and average uncertainty of all words or sequences. 
Results at both the word and sequence levels are illustrated in Fig.~\ref{fig: correlation}. 
Irrelevant words contribute significantly to the overall uncertainty. 
At the sequence level, both the mean and sum uncertainty of irrelevant sequences are prominent. 

Given the substantial proportion of irrelevant words and textual sequences within the generated set, this can introduce unexpected biases and instability when measuring the uncertainty of LLMs-generated answers in real-world open-ended medical QA applications. 
To address these issues, we propose a novel UQ method in the following text. 

\subsection{Word-Sequence Entropy}
In light of the observed issues arising from generative inequality, as demonstrated in Section~\ref{sec: verification}, we propose to emphasize keywords and more semantically relevant responses within the candidate set when conducting UQ. 
To maintain coherence and consistency in the presentation, we strictly adhere to the symbol conventions utilized in Sections~\ref{sec: preliminaries} and~\ref{sec: generative inequality}.

\subsubsection{Word-level WSE}
Since the semantic information carried by each word (token) differs, treating all tokens equally, as described in Eq.~\eqref{eq: sequence-wise entropy 1}, will lead to biased measurements of the predictive uncertainty within each textual sequence. 
To address this, we highlight tokens in keywords by directly multiplying the token-wise entropy by the token-level semantic relevance score: 
\begin{equation} \label{eq: reweighted token-wise entropy}
{\textit{U}}_{T}\left ( {z}_{ijk} \right )={\textit{E}}_{T}\left ( {z}_{ijk} \right ){\textit{R}}_{T}\left ( {z}_{ijk} \right ),
\end{equation}
where ${\textit{E}}_{T}\left ( {z}_{ijk}\right )$ refers to the token-wise entropy of the $k$-th token in the $j$-th word within the $i$-th response. 
Then, the calibrated word-wise entropy of $w_{ij}$ can be formulated as:
\begin{equation} \label{eq: reweighted word-wise entropy} 
{\textit{U}}_{W}\left ( {w}_{ij} \right )=\displaystyle\sum_{k}^{{M}_{j}}{\textit{U}}_{T}\left ( {z}_{ijk} \right ).
\end{equation}
To quantify the overall uncertainty of $\textbf{s}_i$, we sum the calibrated (weighted) uncertainty of all words within this textual sequence:
\begin{equation}\label{eq: reweighted sequence-wise entropy}
{\textit{U}}_{S}\left ( {\textbf{s}}_{i} \right )=\displaystyle\sum_{j}^{{N}_{i}}{\textit{U}}_{W}\left ( {w}_{ij} \right ). 
\end{equation}
Finally, the word-level $\textit{WSE}$ is defined as the arithmetic mean uncertainty of these $K$ candidate responses, following \textit{PE}:
\begin{equation} \label{eq: reweighted word-level uncertainty}
{\textit{WSE}}_{W}\left (\mathbb{S} \right )=\frac{1}{K}\displaystyle\sum_{i}^{K}{U}_{S}\left ( {\textbf{s}}_{i} \right ).
\end{equation}

By employing the word-level $\textit{WSE}$, we capture and highlight keywords carrying the main semantic information within the current textual sequence, thereby calibrating the predictive uncertainty of each candidate response. 

\subsubsection{Sequence-level WSE}
As noted in Section~\ref{sec: generative inequality}, responses, which are semantically consistent with others in the set of $K$ candidate responses, are more trustworthy. 
We reduce the uncertainty associated with the $i$-th textual sequence by adding its generative probability to its semantic relevance score, after dividing by a constant $d$, to obtain the calibrated sequence-wise entropy of ${\textbf{s}}_{i}$:
\begin{equation}\label{eq: reweighted sequence-wise entropy 2}
{{\textit{U}}_{S}}^{\prime}\left ( {\textbf{s}}_{i} \right ) =-\log \left (p\left ( {\textbf{s}}_{i} \right )+\frac{{\textit{R}}_{S}\left ( {\textbf{s}}_{i}\right )}{d}\right ),
\end{equation}
where $d$ serves to regulate the extent to which the semantic relevance score influences the generative probability. 
Operating in the same way as Eq.~\eqref{eq: reweighted word-level uncertainty}, the sequence-level $\textit{WSE}$ is formulated as:
\begin{equation}\label{eq: reweighted sequence-level uncertainty}
{\textit{WSE}}_{S}\left (\mathbb{S} \right )=\frac{1}{K}\displaystyle\sum_{i}^{K}{{\textit{U}}_{S}}^{\prime}\left ( {\textbf{s}}_{i} \right ). 
\end{equation}

By employing the sequence-level \textit{WSE}, we enlarge the generative probability of more reliable responses by assessing their semantic relevance, thereby calibrating the overall uncertainty of the current QA process. 

\subsubsection{Integrated WSE}
Given the direct mathematical relation between the probability and entropy of ${\textbf{s}}_{i}$, as defined in Eq.~\eqref{eq: sequence-wise entropy 2}, we replace $p\left ({\textbf{s}}_{i} \right )$ in Eq.~\eqref{eq: reweighted sequence-wise entropy 2} with ${e}^{-{\textit{U}}_{S}\left ( {\textbf{s}}_{i} \right )}$, where ${\textit{U}}_{S}\left ( {\textbf{s}}_{i} \right )$ represents the calibrated uncertainty of ${\textbf{s}}_{i}$ described in Eq.~\eqref{eq: reweighted sequence-wise entropy}. 
Since the sequence-level semantic relevance score of ${\textbf{s}}_{i}$ (i.e., ${\textit{R}}_{S}\left ( {\textbf{s}}_{i} \right )$) is determined by the probability of compared sequences, we replace $p\left ( {\textbf{s}}_{l} \right  )$ as defined in Eq.~\eqref{eq: sequence-level semantic relevance score}. 
Then, the combined $\textit{WSE}$, which calibrates the uncertainty at both the word and sequence levels, is formulated as: 
\begin{equation}\label{eq: reweighted uncertainty}
{\textit{WSE}}_{C}\left ( \mathbb{S} \right )=\frac{1}{K}\displaystyle\sum_{i}^{K}-\log \left ({{p}_{i}}^{\prime}+\frac{\textstyle\sum_{l\neq i}^{K}{S}_{li}{{p}_{l}}^{\prime}}{d}\right ),
\end{equation}
where ${{p}_{i}}^{\prime}$ and ${{p}_{l}}^{\prime}$ refer to the replacements of the generative probability, and ${S}_{li}$ represents the semantic textual similarity between ${s}_{l}$ and ${s}_{i}$. 
Moreover, The pseudocode of the combined \textit{WSE} is summarized in Algorithm~\ref{alg: algorithm}. 

\input{algorithm/combined_WSE}

In terms of computational complexity, we first analyze \textit{PE}. 
As described in Eqs.~\eqref{eq: sequence-wise entropy 1}-\eqref{eq: predictive entropy}, its computational complexity is $\mathcal{O}\left ( T + K \right )=\mathcal{O}\left ( \max \left ( T, K \right )\right )$, where $T$ is the number of tokens within the response.  
For \textit{SAR}, since it assesses the relevance of each token and sentence within the $K$ generated responses, its computational complexity is $\mathcal{O}\left ( KT+K^2\right )=\mathcal{O}\left (\max \left ( T, K\right) K \right)$. 
Here, $KT$ refers to assessing the relevance of individual tokens within these $K$ responses, and $K^2$ refers to analyzing the similarity between every pair of responses. 
\textit{WSE} assesses semantic relevance as both the word and sequence levels, with the computational complexity of $\mathcal{O}\left ( KN+K^2\right )=\mathcal{O}\left (\max \left ( N, K\right) K \right)$, where $N$ is the number of words within the response, and $KN$ refers to measuring the semantic variation of the $K$ responses before and after removing each word. 
Since a single word can be composed of multiple tokens (i.e., $N \leq T$), \textit{WSE} has a lower computational complexity compared to \textit{SAR}.

By strategically calibrating the uncertainty proportion of keywords and elevating the generative probability of semantically analogous (i.e., more trustworthy) responses, \textit{WSE} focuses more on significant words and responses when estimating the uncertainty of free-form answers generated by LLMs, effectively mitigating biases caused by generative inequality. 
In the latter part of the experiments, we denote word-level $\textit{WSE}$, sequence-level $\textit{WSE}$, and combined $\textit{WSE}$ by ${\textit{WSE}}_{W}$, ${\textit{WSE}}_{S}$, and ${\textit{WSE}}_{C}$, respectively.

%% file: algorithm/combined_WSE.tex
\begin{algorithm}[!t]
\caption{The pseudo-code for the combined $\textit{WSE}$.}\label{alg: algorithm}
\LinesNumbered
\KwIn{$\mathbb{S}$, ${\textbf{s}}_{i}$, ${\textbf{s}}_{l \ne i}$, ${w}_{ij}$, ${z}_{ijk}$, $K$, ${N}_{i}$, ${M}_{j}$, $p\left ({z}_{ijk}\right)$, $d$.}
\For{$i \leftarrow 1$ \KwTo $K$}{
    \For{$j \leftarrow 1$ \KwTo ${N}_{i}$}{
        Calculate the semantic textual similarity of ${\textbf{s}}_{i}$ before and after removing ${w}_{ij}$ $\leftarrow$ ${S}_{W}\left ( {w}_{ij} \right )$;\\
        ${\textit{R}}_{W}\left ( {w}_{ij} \right ) \leftarrow 1 - {S}_{W}\left ( {w}_{ij} \right )$;\\
        \For{$k \leftarrow 1$ \KwTo ${M}_{j}$}{
            ${\textit{R}}_{T}\left ( {z}_{ijk} \right ) \leftarrow {\textit{R}}_{W}\left ( {w}_{ij} \right )$; $\triangleright \; \mathit{consistent} \: \mathit{semantic} \: \mathit{relevance}$\\
            ${\textit{E}}_{T}\left ( {z}_{ijk}\right ) \leftarrow -\log p\left ({z}_{ijk}\right)$;\\
            ${\textit{U}}_{T}\left ( {z}_{ijk}\right ) \leftarrow {\textit{E}}_{T}\left ( {z}_{ijk}\right ){\textit{R}}_{T}\left ( {z}_{ijk} \right )$.
        }
        ${\textit{U}}_{W}\left ( {w}_{ij} \right ) \leftarrow \textstyle\sum_{k}^{{M}_{j}}{\textit{U}}_{T}\left ( {z}_{ijk} \right )$. $\triangleright \; \mathit{word}$-$\mathit{level}$
    }
    ${\textit{U}}_{S}\left ( {s}_{i}\right ) \leftarrow \textstyle\sum_{j}^{{N}_{i}}{\textit{U}}_{W}\left ( {w}_{ij} \right )$; \\
    $p\left ( {s}_{i}\right ) \leftarrow {e}^{-{\textit{U}}_{S}\left ( {s}_{i}\right )}$. $\triangleright \; \mathit{calibrated} \: \mathit{generative} \: \mathit{probability}$
}
\For{$i \leftarrow 1$ \KwTo $K$}{
    \For{$l \leftarrow 1$ \KwTo $K$}{
        \If{$l \ne i$}{
            Calculate the semantic textual similarity between ${s}_{l}$ and ${s}_{i}$ $\leftarrow {S}_{S}\left ( {s}_{l}, {s}_{i} \right )$.
        }
    }
    ${\textit{R}}_{S}\left ( {s}_{i} \right ) \leftarrow \textstyle\sum_{l \neq i}^{K} {S}_{S}\left ( {s}_{l}, {s}_{i} \right )p\left ( {s}_{l}\right  )$;\\
    ${{\textit{U}}_{S}}^{\prime}\left ( {s}_{i} \right ) \leftarrow -\log \left (p\left ( {s}_{i} \right )+\frac{{\textit{R}}_{S}\left ( {s}_{i} \right )}{d}\right )$. $\triangleright \; \mathit{sequence}$-$\mathit{level}$
}
${\textit{WSE}}_{C}\left (\mathbb{S} \right ) \leftarrow \frac{1}{K}\textstyle\sum_{i}^{K} {{\textit{U}}_{S}}^{\prime}\left ( {s}_{i} \right )$.\\
\KwOut{Calibrated predictive uncertainty ${\textit{WSE}}_{C}\left (\mathbb{S} \right )$.}
\end{algorithm}

%% file: section/experiment.tex
\section{Experiments}
\label{sec: experiment}
In this section, we evaluate the performance of $\textit{WSE}$ in accurately measuring the uncertainty of LLMs-generated answers in open-ended medical QA tasks. 
Given the potential for real-world healthcare applications, we resample responses by employing the generation with the lowest uncertainty within the candidate set, measured by $\textit{WSE}$, as the final output to the current medical query, and investigate the overall enhancement of model accuracy. 

\subsection{Experiment Setup}
\subsubsection{Performance Evaluation}
Following Semantic Entropy ($\textit{SE}$)~\cite{kuhn2023semantic,quevedo2024detecting} and \textit{SAR}~\cite{duan2023shifting}, we evaluate \textit{WSE} by framing UQ as the problem of predicting whether to trust a model generation for a given medical query. 
We employ the widely used area under the receiver operating characteristic (AUROC) curve for the binary event that a given response is incorrect, which captures both precision and recall, ranging from 0 to 1, with 1 representing a perfect classifier and 0.5 representing a random estimator. 
This metric evaluates whether \textit{WSE} can effectively distinguish between correct and incorrect answers across various uncertainty thresholds. 
Additionally, since there can be unrealistic uncertainty thresholds, we employ deep AUROC~\cite{carrington2021deep}, which measures performance in multiple groups of predicted risk, or groups of true positive rate or false positive rate.

\subsubsection{Correctness Evaluation}
\label{sec: correctness metric}
We adopt two standard metrics to evaluate the correctness of responses: Rouge-L Similarity (RS)~\cite{lin2004rouge} and Sentence Similarity (SS)~\cite{duan2023shifting}. 
RS measures the longest common subsequence between the output and reference answer, serving as a fuzzy matching criterion. 
For SS, we utilize the cross-encoder model mentioned in Section~\ref{sec: relevance}, with DistillRoBERTa~\cite{sanh2019distilbert} as the backbone. 
SS corresponds to the semantic textual similarity denoted by ${S}_{C}$ in Eq.~\eqref{eq: semantic similarity score}. 
We consider the generation correct if either the RS or SS exceeds the predefined threshold of 0.5. 
Notably, we employ the most likely generation, as introduced in Section~\ref{sec: hyper}, as the object to evaluate the correctness of the current QA. 
In Section~\ref{sec: sensitivity analysis}, we will analyze the sensitivity of \textit{WSE} to these threshold values. 

\subsubsection{Model}
We conduct experiments on seven open-source ``off-the-shelf'' LLMs provided by the Hugging Face platform, including both pre-trained LLMs (e.g., LLaMA-7B~\cite{touvron2023llama}) and instruction-tuned LLMs (e.g., LLaMA-2-7B-Chat~\cite{touvron2023llama2}, Mistral-v0.1~\cite{jiang2023mistral}, Zephyr-7B-Alpha~\cite{tunstall2023zephyr}, Vicuna-7B-v1.5~\cite{zheng2023judging}, WizardLM-7B~\cite{xu2023wizardlm}, StableBeluga-7B~\cite{touvron2023llama2,mukherjee2023orca}) with the model size of 7B. 

\subsubsection{Datasets}
We utilize five free-form medical QA datasets: COVID-QA~\cite{moller2020covid}, Medical Meadow MedQA~\cite{jin2020disease}, PubMedQA~\cite{jin2019pubmedqa}, MedMCQA~\cite{pal2022medmcqa} and MedQuAD~\cite{ben2019question}. 
COVID-QA consists of 2,019 query-answer pairs related to COVID-19, and we employ all query-answer pairs within the maximum sequence length allowed by the language model. 
MedMCQA is a large-scale, multiple-choice QA dataset for medical entrance exams, and we select all samples where a question has only one correct option and begins with ``what'' or ``which'' (1895 in total). 
Medical Meadow MedQA is a free-form multiple-choice OpenQA dataset for solving medical problems, collected from the professional medical board exams. 
MedQuAD covers 37 question types associated with diseases, drugs, and other medical entities such as tests. 
For Medical Meadow MedQA and MedQuAD, we randomly select 2000 test samples from the validation set. 
PubMedQA is a novel biomedical QA dataset collected from PubMed abstracts and we employ the full test set (1000 question-answer pairs).

Unlike COVID-QA and PubMedQA, Medical Meadow MedQA, MedMCQA, and MedQuAD do not provide contextual information, and we randomly select five fixed query-answer pairs from each dataset to form the few-shot prompts, enabling LLMs to follow the instructions.

\subsubsection{Baselines}
We compare our method with $\textit{PE}$~\cite{kadavath2022language}, Semantic Entropy (\textit{SE})~\cite{kuhn2023semantic}, Lexical Similarity ($\textit{LS}$)~\cite{lin2022towards}, Token-level \textit{SAR} (Token-$\textit{SAR}$), Sentence-level \textit{SAR} (Sent-$\textit{SAR}$), and \textit{SAR}~\cite{duan2023shifting}. 
\textit{PE} quantifies uncertainty as described in Section~\ref{sec: generative inequality}. 
\textit{SE} considers semantic equivalence and calculates the cluster-wise entropy. 
\textit{LS} computes the mean semantic similarity score of responses in the set of $K$ generated responses. 
Token-$\textit{SAR}$ and Sent-$\textit{SAR}$ reallocate the uncertainty weights of tokens and sentences based on their relevance, respectively. 
$\textit{SAR}$ combines Token-$\textit{SAR}$ and Sent-$\textit{SAR}$. 
For domain-specific medical QA, ${\textit{WSE}}_{W}$ highlights keywords in each response by assessing the semantic relevance of each word based on semantic variation, which addresses the issue of inconsistent semantic relevance within individual words that \textit{SAR} encounters. 
Additionally, ${\textit{WSE}}_{S}$ leverages a more reliable and explainable measure for semantic textual similarity and enlarges the generative probability of more trustworthy responses based on self-consistency. 
Similar to \textit{SAR}, ${\textit{WSE}}_{C}$ is an orthogonal combination of ${\textit{WSE}}_{W}$ and ${\textit{WSE}}_{S}$, which calibrates uncertainty at both the word and sequence levels.

\input{table/auroc_sentence_similarity}

\subsubsection{Hyperparameters}
\label{sec: hyper}
Given each medical query, LLMs generate five free-form responses (i.e., $K=5$) via multinomial sampling, which are then employed for UQ. 
For the correctness evaluation of the current QA, we employ greedy search to obtain the most likely generation~\cite{kuhn2023semantic, duan2023shifting}. 
The temperature is fixed at 0.5 for all LLMs, and the max length of each generation is set to 128 tokens. 
The coefficient $c$ in Eq.~\eqref{eq: semantic similarity score} is set to 1.0 by default, and the denominator $d$ in the relevance-controlled quantity in Eq.~\eqref{eq: reweighted uncertainty} is empirically set to 0.001. 

\input{table/auroc_rouge}

\subsection{Empirical Findings}
\label{sec: empirical findings}
\subsubsection{Uncertainty Estimation}
\label{sec: comparison}
Given the integrated measurement of semantic textual similarity described in Section~\ref{sec: relevance}, we compare ${\textit{WSE}}_{W}$, ${\textit{WSE}}_{S}$, and ${\textit{WSE}}_{C}$ with six baseline methods, utilizing SS as the criterion for correctness evaluation. 
As summarized in Table~\ref{tb: auroc ss}, all the three \textit{WSE} variants outperform the baseline methods significantly, with ${\textit{WSE}}_{C}$ achieving the highest overall AUROC of 0.6235. 
By highlighting keywords and addressing the inconsistency of semantic relevance within each word, ${\textit{WSE}}_{W}$ surpasses Token-\textit{SAR} by $2.3\%$ in AUROC overall, particularly on the MedQA dataset, where it exceeds Token-\textit{SAR} by $3.17\%$. 
By employing a more reliable measure of semantic similarity, ${\textit{WSE}}_{S}$ surpasses Sent-$\textit{SAR}$ by $1.32\%$ in AUROC overall, especially on the MedMCQA dataset, where it exceeds Sent-\textit{SAR} by $2.53\%$. 
These enhancements highlight the superior adaptability of \textit{WSE} for open-ended medical QA.

In the MedQuAD task, each few-shot prompt comprises multiple question-answer pairs with a similar structure, without providing any contextual information to the language models. 
Additionally, ground truth and generated responses exhibit notably greater length than other tasks. 
To address these challenges, we calculate normalized semantic relevance scores at the word level and assign them to individual tokens. 
This strategy enhances the connectivity between each word and the entire sequence, effectively mitigating biases induced by sequence length.
As a result, ${\textit{WSE}}_{W}$ achieves the highest AUROC of 0.626, significantly outperforming six baseline methods. 

\input{table/deeproc}

Given that RS depends on the length of the longest common subsequence, and semantically equivalent textual sequences can be syntactically or lexically distinct, tasks involving long reference answers and responses may result in no generations meeting the correctness criterion.  
Table~\ref{tb: auroc rouge} presents the comparative results, excluding tasks with an accuracy of 0 from our analysis. 
Despite the inherent evaluation limitations of RS, ${\textit{WSE}}_{W}$ and ${\textit{WSE}}_{C}$ demonstrate remarkable superiority. 
By assessing semantic relevance at the word level rather than evaluating individual tokens independently, ${\textit{WSE}}_{W}$ achieves the second-highest average AUROC of 0.6498, outperforming Token-\textit{SAR} by $2.24\%$, while ${\textit{WSE}}_{C}$ attains the highest average AUROC of 0.6555. 
Notably, comparable baselines exhibit unstable uncertainty estimation under rigorous correctness evaluation conditions (e.g., Sent-\textit{SAR} obtains an AUROC of 0.3647 on the PubMedQA dataset in the Vicuna-7B-v1.5 setting), while \textit{WSE} consistently performs reliably, indicating significant potential for practical medical QA applications in the domain of healthcare.

We also evaluate \textit{WSE} on the COVID-QA dataset utilizing a more stringent deep AUROC metric and correctness evaluation criteria. 
As shown in Table~\ref{tb: deep roc}, all the three variants of \textit{WSE} consistently outperform the corresponding three variants of \textit{SAR}. 
For instance, ${\textit{WSE}}_{S}$ outperforms Sent-\textit{SAR} by $3.2\%$ in deep AUROC at the RS setting, ${\textit{WSE}}_{W}$ surpasses Token-\textit{SAR} by $12.06\%$ at the SS setting, and 
${\textit{WSE}}_{C}$ achieves the highest deep AUROC of 0.7115, exceeding \textit{SAR} by $10.04\%$, with SS as the correctness metric.

Overall, $\textit{WSE}$ demonstrates superior accuracy and stability in quantifying the uncertainty of LLMs-generated responses compared to six baseline methods, utilizing both RS and SS as correctness evaluation criteria across five popular open-ended medical QA tasks. 

\subsubsection{Sensitivity Analysis}
\label{sec: sensitivity analysis}
\begin{figure}[!t]
\centerline{\includegraphics[width=0.6\columnwidth]{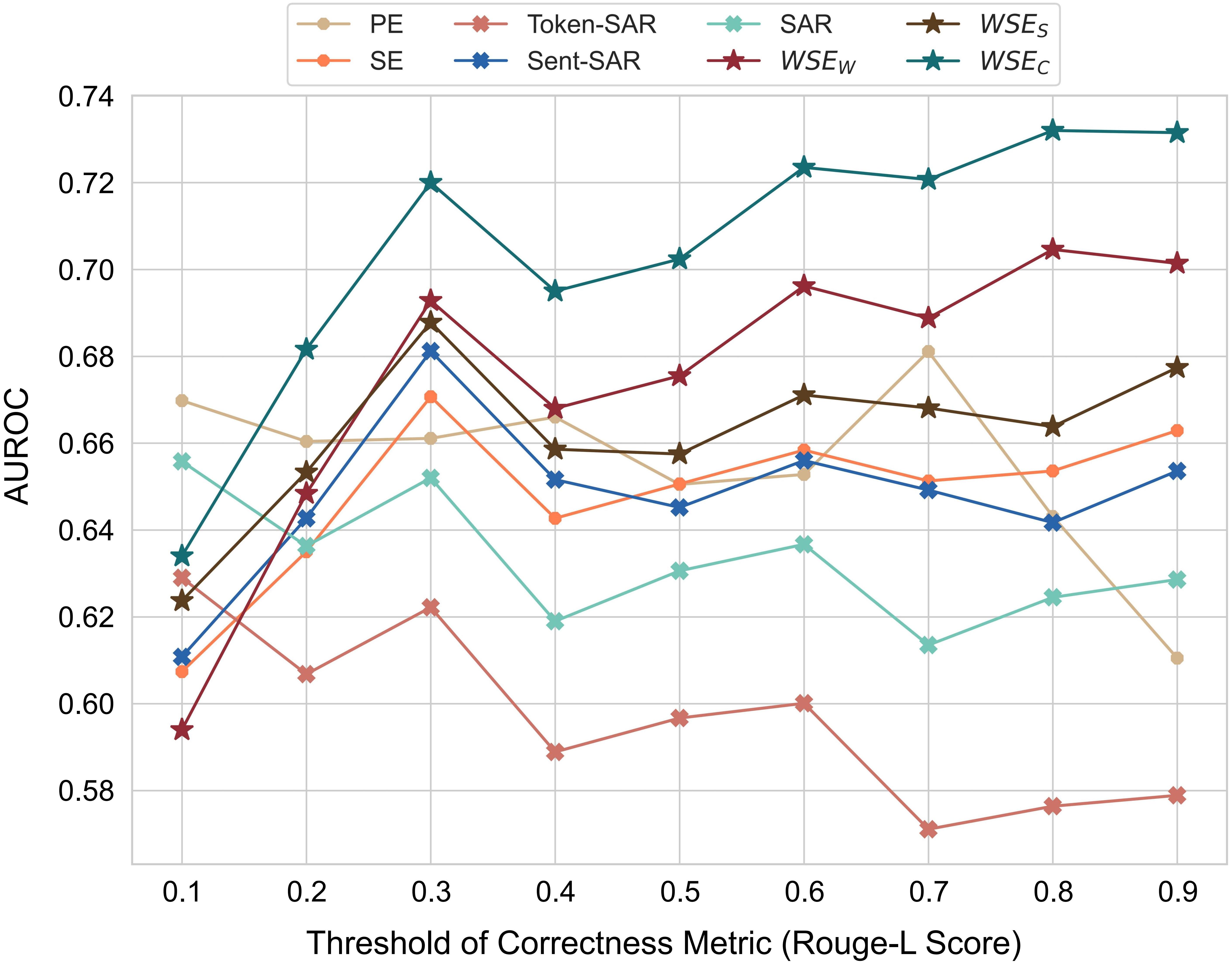}}
\caption{The performance of $\textit{WSE}_W$, $\textit{WSE}_S$, $\textit{WSE}_C$, and five baseline methods at different thresholds of RS. Results are obtained on the COVID-QA dataset utilizing the LLaMA-2-7B-Chat model.}
\label{fig: sensitivity threshold rouge}
\end{figure}

\begin{figure}[!t]
\centerline{\includegraphics[width=0.6\columnwidth]{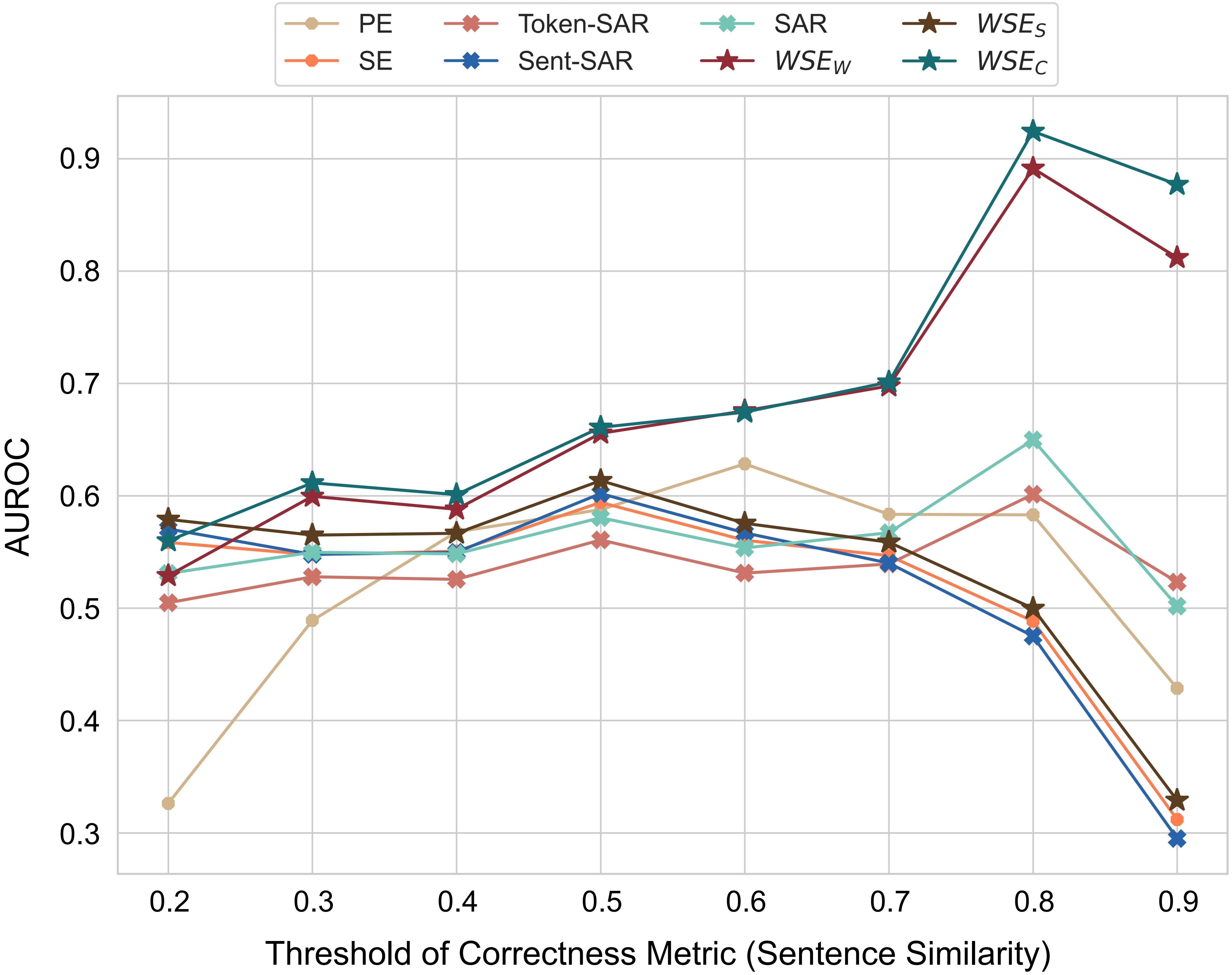}}
\caption{The performance of $\textit{WSE}_W$, $\textit{WSE}_S$, $\textit{WSE}_C$, and five baseline methods at different thresholds of SS. Results are obtained on the COVID-QA dataset utilizing the LLaMA-2-7B-Chat model.}
\label{fig: sensitivity threshold ss}
\end{figure}

To investigate the impact of various thresholds for two correctness metrics on ${\textit{WSE}}_{W}$, ${\textit{WSE}}_{S}$, ${\textit{WSE}}_{C}$, and five baseline methods, we utilize LLaMA-2-7B-Chat and generate ten responses (i.e., $K=10$) to each medical query on the COVID-QA dataset. 
As is shown in Fig.~\ref{fig: sensitivity threshold rouge} and Fig.~\ref{fig: sensitivity threshold ss}, each uncertainty measure is influenced to varying degrees by the threshold. 
Generally, as the evaluation criteria become more stringent, \textit{WSE} consistently outperforms five baseline methods. 
Notably, when utilizing RS, ${\textit{WSE}}_{C}$ achieves the highest AUROC of 0.7315, while using SS results in an AUROC of 0.877. 
When the threshold for SS is set to 0.1, all answers are identified as correct, and we exclude this scenario from our analysis. 

\begin{figure}[!t]
\centerline{\includegraphics[width=0.6\columnwidth]{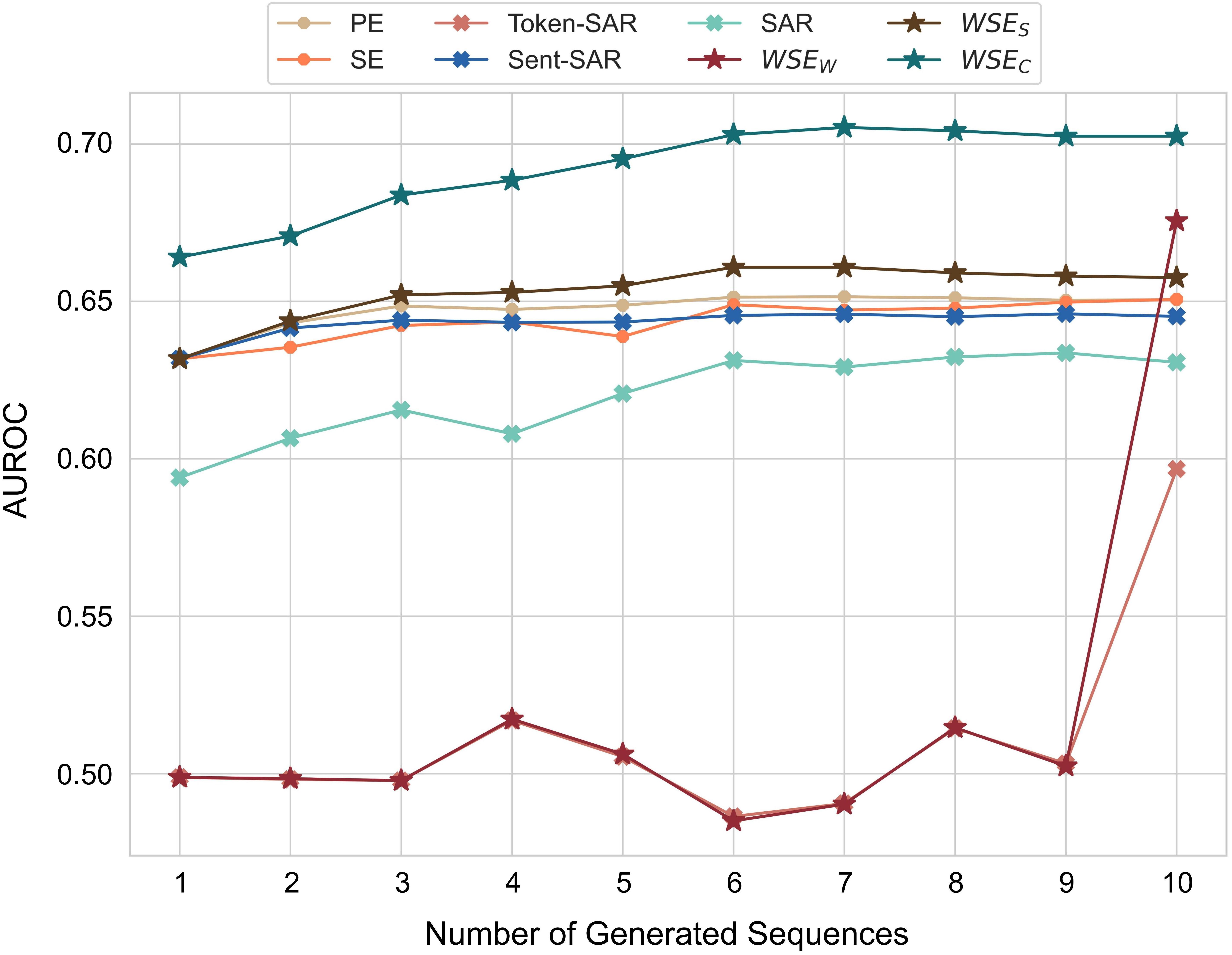}}
\caption{The performance of $\textit{WSE}_W$, $\textit{WSE}_S$, $\textit{WSE}_C$, and baselines at different numbers of generated sequences employing RS as the metric of correctness evaluation. Results are obtained on the COVID-QA dataset utilizing the LLaMA-2-7B-Chat model.}
\label{fig: sensitivity sequences rouge}
\end{figure}

\begin{figure}[!t]
\centerline{\includegraphics[width=0.6\columnwidth]{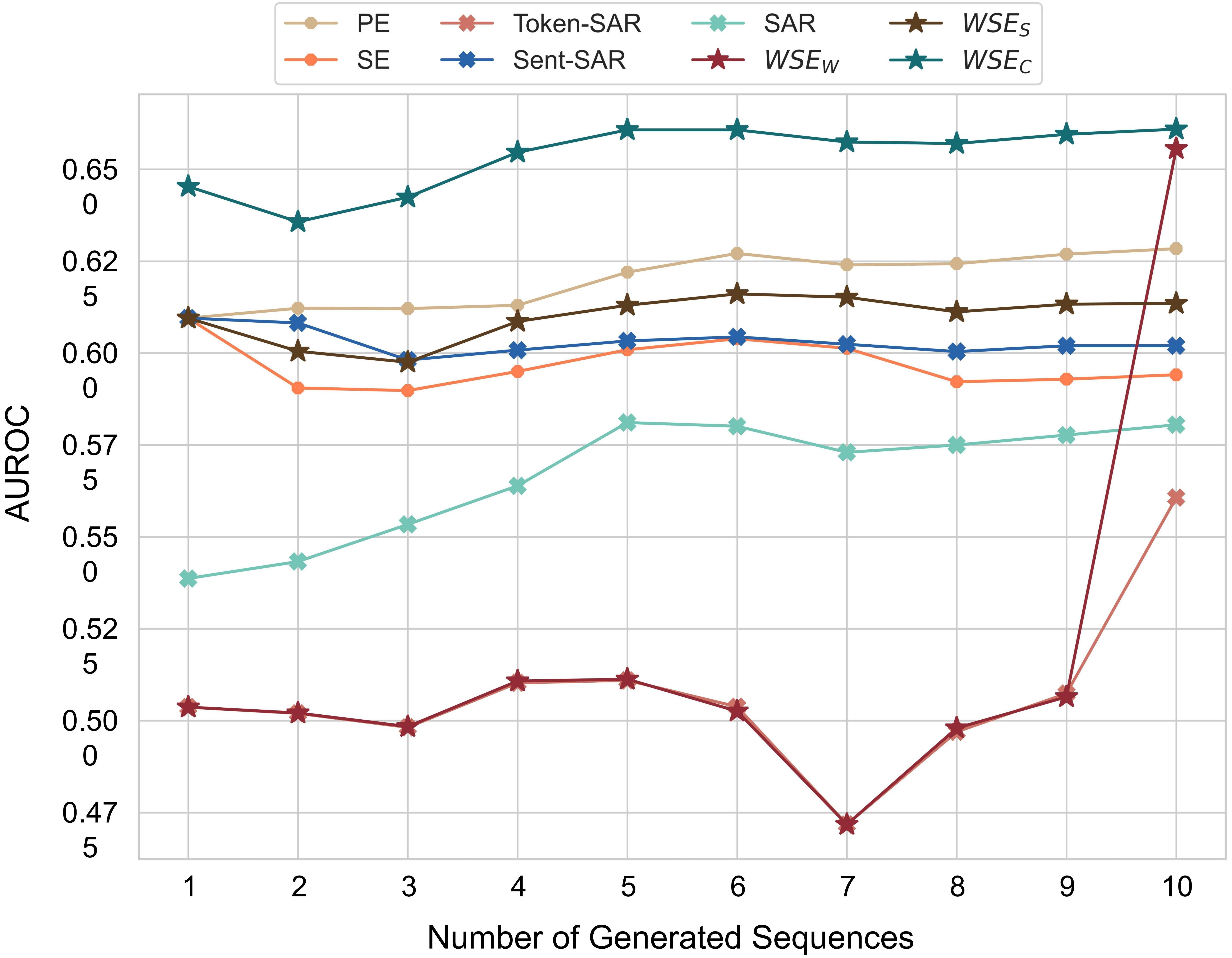}}
\caption{The performance of $\textit{WSE}_W$, $\textit{WSE}_S$, $\textit{WSE}_C$, and baselines at different numbers of generated sequences employing SS as the metric of correctness evaluation. Results are obtained on the COVID-QA dataset utilizing the LLaMA-2-7B-Chat model.}
\label{fig: sensitivity sequences ss}
\end{figure}

Given that entropy-based methods integrate responses within the candidate set, we explore how the number of responses (i.e., $K$) impacts the performance of UQ. 
As illustrated in Fig.~\ref{fig: sensitivity sequences rouge} and Fig.~\ref{fig: sensitivity sequences ss}, ${\textit{WSE}}_{W}$ and Token-\textit{SAR} exhibit sensitivity to variations in $K$. 
Nevertheless, ${\textit{WSE}}_{W}$ ultimately surpasses the baselines and achieves the second-highest AUROC score under both correctness evaluation criteria. 
When employing RS as the correctness metric, ${\textit{WSE}}_{S}$ generally outperforms the baseline methods and achieves the second-highest AUROC of 0.6161 leveraging only 6 generated sequences, which is generation-efficient,. 
It is noteworthy that ${\textit{WSE}}_{C}$ consistently outperforms comparable methods under both correctness evaluation criteria.

\subsection{Accuracy Enhancement}
\label{sec: accuracy}
Due to the abundant and diverse domain-specific knowledge within the healthcare domain, the availability of LLMs specifically designed for open-ended medical QA tasks is comparatively limited. 
Furthermore, real-world medical QA scenarios tend to be highly intricate and often lack contextual information associated with the questions, posing significant challenges to LLMs. 
In this section, we investigate the enhancement of model accuracy solely through resampling and post-processing, by leveraging multiple ``off-the-shelf'' LLMs pre-trained on NLG datasets without requiring additional task-specific training or architectural modifications. 

\input{table/calibrate_accuracy}

Given that \textit{WSE} can quantify uncertainty at the sequence level, we assess the set of $K$ candidate responses, and select the response with the lowest uncertainty, identified by \textit{WSE}, as the final answer to the current medical query. 
Then, we recompute the overall accuracy of the dataset. 

We employ COVID-QA as the dataset and investigate accuracy enhancement under two correctness evaluation criteria. 
As summarized in Table~\ref{tb: calibrate accuracy}, accuracy improvement varies across multiple LLMs when utilizing RS. 
Given that RS is sensitive to the structure of generated sequences, LLMs of the LLaMA series achieve higher initial accuracy than others under two thresholds, with a maximum increase of 6.37$\%$ observed on the LLaMA-2-7B-Chat model. 
After filtering high-uncertainty sequences identified by ${\textit{WSE}}_{W}$, we achieve a substantial accuracy enhancement of 44.56$\%$ on the Mistral model when the correctness metric threshold is set to 0.3. 
Despite the stringent nature and limitations associated with RS, the COVID-QA task exhibits a noteworthy improvement in accuracy across seven ``off-the-shelf'' LLMs. 

For SS, we adopt two relatively stringent thresholds: 0.5 and 0.7. 
Compared to RS, there is no remarkable enhancement in accuracy, with the highest improvement observed at 6.36$\%$ on the Zephyr-7B-Alpha model when the threshold is set to 0.5. 
Overall, the COVID-QA dataset consistently maintains stable and highly effective accuracy improvements, showcasing the significant potential of \textit{WSE} in practical medical QA applications within the domain of healthcare engineering.

%% file: table/auroc_sentence_similarity.tex
\begin{table}[t!]
\centering
\caption{Comparison of ${\textit{WSE}}_{W}$, ${\textit{WSE}}_{S}$, ${\textit{WSE}}_{C}$, and six baseline methods utilizing seven pre-trained and instruction-tuned LLMs on five free-form medical QA datasets, employing SS as the criterion for correctness evaluation with the threshold set to 0.5 (AUROC).}
\adjustbox{max width=\linewidth}{
    \begin{tabular}{c|c|cccccc|ccc}
        \toprule
        Datasets & LLMs & \textit{LS} & \textit{PE} & \textit{SE} & Token-\textit{SAR} & Sent-\textit{SAR} & \textit{SAR} & ${\textit{WSE}}_{W}$ & ${\textit{WSE}}_{S}$ & ${\textit{WSE}}_{C}$\\
        \midrule
        \multirow{7}{*}{COVID-QA} & LLaMA-7B & 0.5076 & 0.7348 & 0.7032 & 0.6903 & 0.7180 & 0.7142 & \underline{0.7448} & 0.7319 & \textbf{0.7454}\\
        \multirow{7}{*}{} & LLaMA-2-7B-Chat & 0.4422 & 0.6756 & 0.6716 & 0.6640 & 0.6765 & 0.6589 & \textbf{0.6869} & 0.6767 & \underline{0.6846}\\
        \multirow{7}{*}{} & Mistral-7B-v0.1 & 0.4341 & 0.7278 & 0.7027 & 0.6911 & 0.7166 & 0.7209 & 0.7318 & \underline{0.7327} & \textbf{0.7482}\\
        \multirow{7}{*}{} & Zephyr-7B-Alpha & 0.4147 & 0.6607 & 0.6583 & 0.6483 & \underline{0.6655} & 0.6558 & 0.6643 & 0.6609 & \textbf{0.6696}\\
        \multirow{7}{*}{} & WizardLM-7B & 0.4059 & 0.6951 & 0.6840 & 0.6737 & 0.6897 & 0.6593 & \textbf{0.7076} & 0.6948 & \underline{0.7016}\\
        \multirow{7}{*}{} & Vicuna-7B-v1.5 & 0.4021 & 0.6955 & 0.6882 & 0.6826 & 0.7011 & 0.6914 & \textbf{0.7159} & 0.6971 & \underline{0.7130}\\
        \multirow{7}{*}{} & StableBeluga-7B & 0.4438 & 0.6904 & 0.7083 & 0.6986 & 0.7027 & 0.6962 & \underline{0.7121} & 0.7068 & \textbf{0.7228}\\

        \midrule
        \multicolumn{2}{c|}{Average} & 0.4358 & 0.6971 & 0.6880 & 0.6784 & 0.6957 & 0.6857 & \underline{0.7091} & 0.7001 & \textbf{0.7122}\\
        
        \midrule
        \multirow{7}{*}{MedQA} & LLaMA-7B & 0.5143 & 0.5122 & \underline{0.5493} & 0.4789 & 0.5468 & 0.5130 & 0.5164 & 0.5438 & \textbf{0.5502}\\
        \multirow{7}{*}{} & LLaMA-2-7B-Chat & 0.5483 & 0.5793 & 0.5958 & 0.5805 & 0.5948 & \underline{0.6145} & 0.6102 & 0.6074 & \textbf{0.6415}\\
        \multirow{7}{*}{} & Mistral-7B-v0.1 & 0.5355 & 0.4845 & 0.5119 & 0.4915 & 0.5085 & \underline{0.5517} & 0.5185 & 0.5506 & \textbf{0.5782}\\
        \multirow{7}{*}{} & Zephyr-7B-Alpha & 0.5035 & 0.4979 & 0.5206 & 0.4936 & 0.5043 & 0.5251 & 0.5192 & \underline{0.5326} & \textbf{0.5619}\\
        \multirow{7}{*}{} & WizardLM-7B & 0.5985 & 0.4631 & 0.5684 & 0.4836 & \underline{0.6286} & 0.5517 & 0.5499 & \textbf{0.6314} & 0.6211\\
        \multirow{7}{*}{} & Vicuna-7B-v1.5 & 0.5079 & 0.4538 & 0.4752 & 0.5093 & 0.4510 & \underline{0.5335} & 0.5295 & 0.4746 & \textbf{0.5576}\\
        \multirow{7}{*}{} & StableBeluga-7B & \textbf{0.5776} & 0.5139 & 0.5481 & 0.5318 & 0.5749 & 0.5696 & 0.5474 & \underline{0.5758} & 0.5749\\

        \midrule
        \multicolumn{2}{c|}{Average} & 0.5408 & 0.5007 & 0.5385 & 0.5099 & 0.5441 & 0.5513 & 0.5416 & \underline{0.5595} & \textbf{0.5836}\\

        \midrule
        \multirow{7}{*}{MedMCQA} & LLaMA-7B & 0.5468 & 0.5290 & 0.5415 & 0.5394 & \underline{0.5583} & 0.5399 & 0.5498 & \textbf{0.5586} & 0.5548\\
        \multirow{7}{*}{} & LLaMA-2-7B-Chat & 0.5108 & 0.4954 & 0.5015 & 0.5128 & 0.4833 & 0.5200 & \underline{0.5467} & 0.5030 & \textbf{0.5612}\\
        \multirow{7}{*}{} & Mistral-7B-v0.1 & 0.5075 & 0.4909 & 0.5216 & 0.5205 & 0.4980 & 0.5523 & 0.5146 & \underline{0.5584} & \textbf{0.5777}\\
        \multirow{7}{*}{} & Zephyr-7B-Alpha & 0.4831 & 0.5175 & 0.5404 & 0.5356 & 0.5331 & 0.5374 & 0.5259 & \textbf{0.5534} & \underline{0.5512}\\
        \multirow{7}{*}{} & WizardLM-7B & 0.5320 & 0.4980 & 0.5074 & 0.4957 & 0.5025 & 0.5063 & \underline{0.5517} & 0.5149 & \textbf{0.5623}\\
        \multirow{7}{*}{} & Vicuna-7B-v1.5 & 0.5016 & 0.4952 & 0.5015 & 0.5011 & 0.4803 & 0.5065 & \underline{0.5288} & 0.4975 & \textbf{0.5395}\\
        \multirow{7}{*}{} & StableBeluga-7B & 0.4990 & 0.4446 & 0.4833 & 0.4446 & 0.4655 & \underline{0.5305} & 0.4421 & 0.5125 & \textbf{0.5314}\\

        \midrule
        \multicolumn{2}{c|}{Average} & 0.5115 & 0.4958 & 0.5139 & 0.5071 & 0.5030 & 0.5276 & 0.5228 & \underline{0.5283} & \textbf{0.5540}\\

        \midrule
        \multirow{7}{*}{PubMedQA} & LLaMA-7B & 0.5496 & 0.5424 & 0.6202 & 0.5414 & 0.6129 & 0.6269 & 0.5420 & \textbf{0.6343} & \underline{0.6340}\\
        \multirow{7}{*}{} & LLaMA-2-7B-Chat & 0.5024 & 0.6146 & 0.5918 & 0.5676 & 0.5819 & 0.6176 & 0.5736 & \underline{0.6179} & \textbf{0.6640}\\
        \multirow{7}{*}{} & Mistral-7B-v0.1 & 0.5018 & 0.6440 & \underline{0.6644} & 0.5262 & 0.6614 & 0.6022 & 0.5808 & \textbf{0.6980} & 0.6627\\
        \multirow{7}{*}{} & Zephyr-7B-Alpha & \underline{0.5929} & 0.5682 & 0.5706 & 0.4894 & 0.5594 & 0.5310 & 0.5293 & 0.5793 & \textbf{0.6027}\\
        \multirow{7}{*}{} & WizardLM-7B & 0.5587 & 0.5308 & 0.5525 & 0.4676 & 0.5265 & 0.5172 & 0.5080 & \underline{0.5640} & \textbf{0.6031}\\
        \multirow{7}{*}{} & Vicuna-7B-v1.5 & 0.5787 & 0.6631 & \underline{0.6728} & 0.5715 & 0.6617 & 0.6289 & 0.6112 & 0.6670 & \textbf{0.6869}\\
        \multirow{7}{*}{} & StableBeluga-7B & 0.6075 & 0.6598 & 0.6461 & 0.6662 & 0.6419 & \underline{0.6971} & 0.6664 & 0.6754 & \textbf{0.7169}\\

        \midrule
        \multicolumn{2}{c|}{Average} & 0.5559 & 0.6033 & 0.6169 & 0.5471 & 0.6065 & 0.603 & 0.5730 & \underline{0.6337} & \textbf{0.6529}\\

        \midrule
        \multirow{7}{*}{MedQuAD} & LLaMA-7B & \underline{0.6546} & 0.5996 & 0.6040 & 0.6534 & 0.6446 & 0.6491 & \textbf{0.6618} & 0.6365 & 0.6502\\
        \multirow{7}{*}{} & LLaMA-2-7B-Chat & 0.5758 & 0.4889 & 0.5123 & 0.5743 & 0.5484 & \underline{0.5884} & 0.5879 & 0.5364 & \textbf{0.5890}\\
        \multirow{7}{*}{} & Mistral-7B-v0.1 & 0.5838 & \textbf{0.6091} & 0.5409 & 0.578 & 0.5639 & 0.5718 & 0.5823 & 0.5643 & \underline{0.5847}\\
        \multirow{7}{*}{} & Zephyr-7B-Alpha & 0.5718 & 0.5012 & 0.6283 & 0.6732 & 0.6393 & 0.6673 & \textbf{0.6817} & 0.6327 & \underline{0.6756}\\
        \multirow{7}{*}{} & WizardLM-7B & 0.5866 & 0.4447 & 0.5405 & 0.5958 & 0.5613 & 0.5871 & \textbf{0.6112} & 0.5596 & \underline{0.6003}\\
        \multirow{7}{*}{} & Vicuna-7B-v1.5 & 0.5748 & 0.4469 & 0.5652 & \underline{0.6357} & 0.5792 & 0.6249 & \textbf{0.6493} & 0.5727 & 0.6301\\
        \multirow{7}{*}{} & StableBeluga-7B & 0.5671 & 0.5226 & 0.5887 & \underline{0.5960} & 0.5738 & 0.5732 & \textbf{0.608} & 0.5634 & 0.5737\\
        
        \midrule
        \multicolumn{2}{c|}{Average} & 0.5878 & 0.5161 & 0.5686 & 0.6152 & 0.5872 & 0.6088 & \textbf{0.6260} & 0.5808 & \underline{0.6148}\\

        \midrule
        \multicolumn{2}{c|}{Overall} & 0.5264 & 0.5626 & 0.5852 & 0.5715 & 0.5873 & 0.5952 & 0.5945 & \underline{0.6005} & \textbf{0.6235}\\
        \bottomrule
    \end{tabular}
}
\label{tb: auroc ss}
\end{table}

%% file: table/auroc_rouge.tex
\begin{table}[t!]
\centering
\caption{Comparison of ${\textit{WSE}}_{W}$, ${\textit{WSE}}_{S}$, ${\textit{WSE}}_{C}$, and six baseline methods utilizing seven pre-trained and instruction-tuned LLMs on five free-form medical QA datasets, employing RS as the criterion for correctness evaluation with the threshold set to 0.5 (AUROC).}
\adjustbox{max width=\linewidth}{
    \begin{tabular}{c|c|cccccc|ccc}
        \toprule
        Datasets & LLMs & \textit{LS} & \textit{PE} & \textit{SE} & Token-\textit{SAR} & Sent-\textit{SAR} & \textit{SAR} & ${\textit{WSE}}_{W}$ & ${\textit{WSE}}_{S}$ & ${\textit{WSE}}_{C}$\\
        \midrule
        \multirow{7}{*}{COVID-QA} & LLaMA-7B & 0.5726 & 0.7297 & 0.7114 & 0.6735 & 0.7159 & 0.7047 & 0.7108 & \underline{0.7304} & \textbf{0.7445}\\
        \multirow{7}{*}{} & LLaMA-2-7B-Chat & 0.4676 & 0.7164 & 0.7148 & 0.7103 & 0.7174 & 0.7098 & \underline{0.7255} & 0.7223 & \textbf{0.7324}\\
        \multirow{7}{*}{} & Mistral-7B-v0.1 & 0.5403 & 0.6368 & 0.6530 & 0.6697 & 0.6367 & 0.6418 & \textbf{0.7207} & 0.6432 & \underline{0.6922}\\
        \multirow{7}{*}{} & Zephyr-7B-Alpha & 0.5748 & 0.6344 & 0.6445 & 0.6015 & 0.6381 & 0.6120 & 0.6416 & \underline{0.6458} & \textbf{0.6524}\\
        \multirow{7}{*}{} & WizardLM-7B & 0.5591 & 0.6455 & \textbf{0.6623} & 0.6401 & 0.6345 & 0.5569 & \underline{0.6486} & 0.6341 & 0.5779\\
        \multirow{7}{*}{} & Vicuna-7B-v1.5 & 0.4898 & 0.6699 & 0.6972 & 0.6981 & 0.6959 & 0.6405 & \underline{0.7051} & 0.6936 & \textbf{0.7528}\\
        \multirow{7}{*}{} & StableBeluga-7B & 0.5795 & 0.6730 & 0.6712 & 0.6647 & 0.6744 & 0.6376 & \textbf{0.6839} & \underline{0.6744} & 0.6528\\
        
        \midrule
        \multirow{7}{*}{MedQA} & LLaMA-7B & 0.5162 & 0.5191 & 0.5620 & 0.4725 & \underline{0.5730} & 0.5178 & 0.5257 & 0.5679 & \textbf{0.5739}\\
        \multirow{7}{*}{} & LLaMA-2-7B-Chat & 0.5714 & 0.5874 & 0.6194 & 0.5666 & 0.6265 & 0.6192 & 0.6167 & \underline{0.6364} & \textbf{0.6581}\\
        \multirow{7}{*}{} & Mistral-7B-v0.1 & 0.5456 & 0.5246 & 0.5358 & 0.4952 & 0.5541 & 0.5706 & 0.5004 & \underline{0.5826} & \textbf{0.5828}\\
        \multirow{7}{*}{} & Zephyr-7B-Alpha & 0.4897 & 0.5278 & 0.5451 & 0.4987 & 0.5327 & 0.5400 & 0.5212 & \underline{0.5633} & \textbf{0.5803}\\
        \multirow{7}{*}{} & WizardLM-7B & 0.6246 & 0.4668 & 0.5755 & 0.4808 & 0.6161 & 0.5565 & 0.5461 & \textbf{0.6285} & \underline{0.6268}\\
        \multirow{7}{*}{} & Vicuna-7B-v1.5 & 0.5154 & 0.4967 & 0.5085 & 0.5128 & 0.4937 & \underline{0.5426} & 0.5274 & 0.5126 & \textbf{0.5608}\\
        \multirow{7}{*}{} & StableBeluga-7B & 0.5860 & 0.5378 & 0.5741 & 0.5522 & 0.6048 & 0.5949 & 0.5687 & \underline{0.6074} & \textbf{0.6097}\\

        \midrule
        \multirow{6}{*}{MedMCQA} & LLaMA-7B & 0.5596 & 0.5182 & 0.5511 & 0.5347 & \underline{0.5693} & 0.5403 & 0.5463 & \textbf{0.5699} & 0.5589\\
        \multirow{6}{*}{} & LLaMA-2-7B-Chat & 0.5030 & 0.4988 & 0.5012 & 0.5347 & 0.4881 & 0.5453 & \underline{0.5544} & 0.5076 & \textbf{0.5720}\\
        \multirow{6}{*}{} & Mistral-7B-v0.1 & 0.5293 & 0.5307 & 0.5382 & 0.5295 & 0.5402 & 0.5652 & 0.5168 & \underline{0.5760} & \textbf{0.5781}\\
        \multirow{6}{*}{} & Zephyr-7B-Alpha & 0.4801 & 0.5500 & 0.5896 & 0.5659 & 0.5842 & 0.5718 & 0.5536 & \textbf{0.6103} & \underline{0.5921}\\
        \multirow{6}{*}{} & WizardLM-7B & 0.5151 & 0.5051 & 0.5000 & 0.5124 & 0.5005 & 0.5268 & \underline{0.5381} & 0.5095 & \textbf{0.5395}\\
        \multirow{6}{*}{} & Vicuna-7B-v1.5 & 0.5048 & 0.4983 & 0.4937 & 0.5182 & 0.4843 & \underline{0.5311} & 0.5304 & 0.5031 & \textbf{0.5499}\\

        \midrule
        \multirow{5}{*}{PubMedQA} & LLaMA-7B & 0.5123 & \underline{0.5457} & 0.5401 & 0.5423 & 0.5203 & 0.5426 & 0.5403 & 0.5451 & \textbf{0.5540}\\
        \multirow{5}{*}{} & LLaMA-2-7B-Chat & 0.6511 & 0.6146 & 0.5867 & 0.6329 & 0.5726 & \underline{0.7053} & 0.6420 & 0.6028 & \textbf{0.7329}\\
        \multirow{5}{*}{} & Mistral-7B-v0.1 & 0.5172 & 0.5331 & 0.5231 & 0.5093 & 0.5131 & 0.5133 & 0.5094 & \underline{0.5654} & \textbf{0.5659}\\
        \multirow{5}{*}{} & Zephyr-7B-Alpha & 0.4945 & 0.6194 & 0.4433 & 0.6103 & 0.4408 & 0.5737 & \textbf{0.6640} & 0.4754 & \underline{0.6545}\\
        \multirow{5}{*}{} & Vicuna-7B-v1.5 & 0.7465 & 0.3397 & 0.3838 & \underline{0.8246} & 0.3647 & 0.7926 & \textbf{0.8888} & 0.3477 & 0.6283\\

        \midrule
        \multirow{5}{*}{MedQuAD} & LLaMA-7B & \underline{0.7442} & 0.7123 & 0.6611 & 0.6821 & 0.7126 & 0.7161 & 0.7216 & 0.7108 & \textbf{0.7470}\\
        \multirow{5}{*}{} & LLaMA-2-7B-Chat & 0.8667 & 0.8783 & 0.8353 & \underline{0.9215} & 0.8413 & 0.9065 & \textbf{0.9527} & 0.8423 & 0.9108\\
        \multirow{5}{*}{} & Mistral-7B-v0.1 & 0.7893 & 0.7985 & 0.6101 & \underline{0.8292} & 0.6700 & 0.8072 & \textbf{0.8394} & 0.7291 & 0.7040\\
        \multirow{5}{*}{} & WizardLM-7B & 0.2327 & 0.9816 & 0.9718 & \underline{0.9843} & 0.9775 & 0.9721 & \textbf{0.9889} & 0.9746 & 0.9561\\
        \multirow{5}{*}{} & Vicuna-7B-v1.5 & 0.9065 & 0.9020 & 0.9014 & 0.9266 & 0.9035 & \underline{0.9275} & \textbf{0.9356} & 0.9125 & 0.9142\\
        
        \midrule
        \multicolumn{2}{c|}{Overall} & 0.5729 & 0.6131 & 0.6102 & 0.6298 & 0.6132 & 0.6394 & \underline{0.6522} & 0.6275 & \textbf{0.6585}\\
        \bottomrule
    \end{tabular}
}
\label{tb: auroc rouge}
\end{table}

%% file: table/deeproc.tex
\begin{table}[t!]
\centering
\caption{Comparison of \textit{WSE} and \textit{SAR} utilizing seven popular LLMs on the COVID-QA datasets, employing both SS and RS as the criteria for correctness evaluation with a more stringent threshold set to 0.7 (deep AUROC).}
\adjustbox{max width=\linewidth}{
    \begin{tabular}{c|c|ccc|ccc}
        \toprule
        Metrics & LLMs & Token-\textit{SAR} & Sent-\textit{SAR} & \textit{SAR} & ${\textit{WSE}}_{W}$ & ${\textit{WSE}}_{S}$ & ${\textit{WSE}}_{C}$\\
        \midrule
        \multirow{7}{*}{RS} & LLaMA-7B  & 0.6846 & 0.7217 & 0.7280 & 0.7194 & \underline{0.7397} & \textbf{0.7423}\\
        \multirow{7}{*}{} & LLaMA-2-7B-Chat & 0.5774 & 0.6501 & 0.6234 & 0.7046 & \textbf{0.7309} & \underline{0.7297}\\
        \multirow{7}{*}{} & Mistral-7B-v0.1 & \underline{0.6486} & 0.6052 & 0.6188 & \textbf{0.6526} & 0.6205 & 0.6352\\
        \multirow{7}{*}{} & Zephyr-7B-Alpha  & 0.5373 & 0.6087 & 0.5120 & 0.6144 & \textbf{0.6253} & \underline{0.6237}\\
        \multirow{7}{*}{} & WizardLM-7B & 0.7645 & 0.7277 & 0.7038 &\textbf{ 0.8092} & 0.7835 & \underline{0.7944}\\
        \multirow{7}{*}{} & Vicuna-7B-v1.5  & 0.6490 & 0.6637 & 0.5336 & \underline{0.6691} & \textbf{0.6842} & 0.6216\\
        \multirow{7}{*}{} & StableBeluga-7B & \underline{0.7942} & 0.6860 & 0.6658 & \textbf{0.8113} & 0.7035 & 0.6703\\

        \midrule
        \multicolumn{2}{c|}{Average} & 0.6651 & 0.6662 & 0.6265 & \textbf{0.7115} & \underline{0.6982} & 0.6882\\
        
        \midrule
        \multirow{7}{*}{SS} & LLaMA-7B & 0.5523 & 0.6775 & 0.5828 & \underline{0.7410} & 0.6871 & \textbf{0.7468}\\
        \multirow{7}{*}{} & LLaMA-2-7B-Chat & 0.5372 & 0.5213 & 0.5703 & \underline{0.6909} & 0.5484 & \textbf{0.6972}\\
        \multirow{7}{*}{} & Mistral-7B-v0.1 & 0.5476 & 0.6643 & 0.5724 & \textbf{0.7556} & 0.6657 & \underline{0.7425}\\
        \multirow{7}{*}{} & Zephyr-7B-Alpha & 0.4927 & 0.6759 & 0.5218 & \underline{0.6983} & 0.6741 & \textbf{0.7211}\\
        \multirow{7}{*}{} & WizardLM-7B & 0.6129 & \underline{0.6584} & 0.6354 & 0.6242 & \textbf{0.6691} & 0.6525\\
        \multirow{7}{*}{} & Vicuna-7B-v1.5 & 0.6508 & 0.6783 & 0.6766 & \underline{0.6979} & 0.6829 & \textbf{0.7010}\\
        \multirow{7}{*}{} & StableBeluga-7B & 0.6996 & 0.6761 & 0.7187 & \textbf{0.7293} & 0.6762 & \underline{0.7191}\\

        \midrule
        \multicolumn{2}{c|}{Average} & 0.5847 & 0.6503 & 0.6111 & \underline{0.7053} & 0.6576 & \textbf{0.7115}\\

        \bottomrule
    \end{tabular}
}
\label{tb: deep roc}
\end{table}

%% file: table/calibrate_accuracy.tex
\begin{table}[t!]
\centering
\renewcommand\arraystretch{1.3}
\caption{The enhancement of model accuracy after employing responses with lower uncertainty identified by ${\textit{WSE}}_{W}$, ${\textit{WSE}}_{S}$, and ${\textit{WSE}}_{C}$, utilizing both RS and SS as the criteria for correctness evaluation under multiple thresholds. Experimental results are obtained on the COVID-QA dataset.}
\adjustbox{max width=\linewidth}{
    \begin{tabular}{c|c|c|c|ccccccc}
        \toprule
        Metrics & Threshold & \multicolumn{2}{c|}{Accuracy} & LLaMA & LLaMA-Chat & Mistral & Zephyr & WizardLM & Vicuna & StabeBeluga\\
        
        \midrule
        
        \multirow{10}{*}{RS}  & \multirow{5}{*}{0.3} & \multicolumn{2}{c|}{Initial} & 0.4775 & 0.5172 & 0.1777 & 0.1936 & 0.1485 & 0.1406 & 0.1777\\
        \cline{3-4}
        {} & \multirow{5}{*}{} & \multirow{3}{*}{Calibrated} & ${\textit{WSE}}_{W}$ & 0.5225 & 0.5809 & 0.6233 & 0.5862 & 0.5517 & 0.5544 & 0.5703\\
        {} & \multirow{5}{*}{} & \multirow{3}{*}{} & ${\textit{WSE}}_{S}$ & 0.5013 & 0.557 & 0.5995 & 0.5597 & 0.496 & 0.5279 & 0.5438\\
        {} & \multirow{5}{*}{} & \multirow{3}{*}{} & ${\textit{WSE}}_{C}$ & 0.504 & 0.557 & 0.5968 & 0.557 & 0.4934 & 0.5279 & 0.5438\\
        \cline{3-11}
        {} & \multirow{5}{*}{} & \multicolumn{2}{c|}{Enhanced (max)} & $\uparrow 4.5\%$ & $\uparrow 6.37\%$ & $\uparrow 44.56\%$ & $\uparrow 39.26\%$ & $\uparrow 40.32\%$ & $\uparrow 41.38\%$ & $\uparrow 39.26\%$\\
        
        \cline{2-11}
        
        {} & \multirow{5}{*}{0.5} & \multicolumn{2}{c|}{Initial} & 0.3475 & 0.3899 & 0.0849 & 0.0769 & 0.0504 & 0.0637 & 0.0557\\
        \cline{3-4}
        {} & \multirow{5}{*}{} & \multirow{3}{*}{Calibrated} & ${\textit{WSE}}_{W}$ & 0.382 & 0.4138 & 0.4881 & 0.4191 & 0.3687 & 0.3979 & 0.3793\\
        {} & \multirow{5}{*}{} & \multirow{3}{*}{} & ${\textit{WSE}}_{S}$ & 0.3501 & 0.4032 & 0.4562 & 0.3395 & 0.313 & 0.3395 & 0.3289\\
        {} & \multirow{5}{*}{} & \multirow{3}{*}{} & ${\textit{WSE}}_{C}$ & 0.3448 & 0.4085 & 0.4562 & 0.3342 & 0.321 & 0.3395 & 0.3236\\
        \cline{3-11}
        {} & \multirow{5}{*}{} & \multicolumn{2}{c|}{Enhanced (max)} & $\uparrow 3.45\%$ & $\uparrow 2.39\%$ & $\uparrow 40.32\%$ & $\uparrow 34.22\%$ & $\uparrow 31.83\%$ & $\uparrow 33.42\%$ & $\uparrow 32.36\%$\\
        
        \midrule

        \multirow{10}{*}{SS}  & \multirow{5}{*}{0.5} & \multicolumn{2}{c|}{Initial} & 0.2679 & 0.2759 & 0.3024 & 0.1910 & 0.2122 & 0.2334 & 0.1857\\
        \cline{3-4}
        {} & \multirow{5}{*}{} & \multirow{3}{*}{Calibrated} & ${\textit{WSE}}_{W}$ & 0.2918 & 0.2865 & 0.3422 & 0.2546 & 0.2202 & 0.2653 & 0.2122\\
        {} & \multirow{5}{*}{} & \multirow{3}{*}{} & ${\textit{WSE}}_{S}$ & 0.2891 & 0.2785 & 0.3263 & 0.2202 & 0.2042 & 0.252 & 0.1963\\
        {} & \multirow{5}{*}{} & \multirow{3}{*}{} & ${\textit{WSE}}_{C}$ & 0.2679 & 0.2706 & 0.3103 & 0.2122 & 0.1989 & 0.2361 & 0.1883\\
        \cline{3-11}
        {} & \multirow{5}{*}{} & \multicolumn{2}{c|}{Enhanced (max)} & $\uparrow 2.39\%$ & $\uparrow 1.06\%$ & $\uparrow 3.98\%$ & $\uparrow 6.36\%$ & $\uparrow 0.8\%$ & $\uparrow 3.19\%$ & $\uparrow 2.65\%$\\
        
        \cline{2-11}
        
        {} & \multirow{5}{*}{0.7} & \multicolumn{2}{c|}{Initial} & 0.1008 & 0.1061 & 0.1273 & 0.0584 & 0.069 & 0.0769 & 0.0584\\
        \cline{3-4}
        {} & \multirow{5}{*}{} & \multirow{3}{*}{Calibrated} & ${\textit{WSE}}_{W}$ & 0.13 & 0.1114 & 0.1485 & 0.1008 & 0.0743 & 0.1008 & 0.069\\
        {} & \multirow{5}{*}{} & \multirow{3}{*}{} & ${\textit{WSE}}_{S}$ & 0.1167 & 0.1008 & 0.1406 & 0.0716 & 0.0743 & 0.0902 & 0.0557\\
        {} & \multirow{5}{*}{} & \multirow{3}{*}{} & ${\textit{WSE}}_{C}$ & 0.1141 & 0.1034 & 0.13 & 0.0637 & 0.0743 & 0.0849 & 0.0769\\
        \cline{3-11}
        {} & \multirow{5}{*}{} & \multicolumn{2}{c|}{Enhanced (max)} & $\uparrow 2.92\%$ & $\uparrow 0.53\%$ & $\uparrow 2.12\%$ & $\uparrow 4.24\%$ & $\uparrow 0.53\%$ & $\uparrow 2.39\%$ & $\uparrow 1.85\%$\\

        \bottomrule
    \end{tabular}
}
\label{tb: calibrate accuracy}
\end{table}

%% file: section/conclusion.tex
\section{Conclusion}
\label{sec: conclusion}
We address the lack of general uncertainty measures in open-ended medical QA tasks. 
Given that generative inequality leads to a large number of irrelevant words and responses in the candidate set for UQ, we highlight the keywords within each textual sequence based on semantic variation and enlarge the generative probability of reliable
responses through self-consistency. 
In the UQ process, we develop a stable measure of semantic textual similarity. 
Furthermore, to overcome the limitations of LLMs in medical QA, we focus on posterior work and utilize sequences with lower uncertainty identified by \textit{WSE} as final answers, significantly enhancing model accuracy. 
Experiments on five medical QA datasets demonstrate the superior performance of \textit{WSE} in accurate UQ and its substantial potential in healthcare. 

Our proposed method employs ``off-the-shelf'' LLMs without requiring additional fine-tuning or modifications (i.e., unsupervised), facilitating further research in this area and enhancing reproducibility. 
However, with the rise of closed-source LLMs served via APIs, end-users typically lack access to token likelihoods or embeddings, limiting the applicability of entropy-based measures. 
A promising future research direction is to explore black-box approaches for estimating the confidence or uncertainty of LLMs in their responses. 
Additionally, the semantic diversity of the model's output space cannot fully capture the nuances of its uncertainty. 
A more comprehensive analysis is warranted, considering factors such as the model's design mechanism and data noise. 
Furthermore, the reliability of semantic similarity scores significantly affects the sensitivity of semantics-based approaches. 
We will investigate measurements of semantic textual similarity with stronger explainability and trustworthiness, and aim to devise certified methods for a theoretically rigorous uncertainty notion. 
By providing users with information regarding the uncertainty of language model outputs, we endeavor to advance the development of safer and more trustworthy QA systems, particularly in the domain of healthcare engineering. 